\documentclass[journal,compsoc]{IEEEtran}

\usepackage[numbers]{natbib}

\usepackage{multirow}

\usepackage[utf8]{inputenc} 
\usepackage[T1]{fontenc}    
\usepackage{hyperref}       
\usepackage{url}            
\usepackage{booktabs}       
\usepackage{amsfonts}       
\usepackage{nicefrac}       
\usepackage{microtype}      
\usepackage{xcolor}         

\usepackage{graphicx}
\usepackage{amsmath}
\usepackage{amssymb}
\usepackage{xcolor}
\usepackage{caption}
\usepackage{subcaption}
\usepackage{mathtools}
\usepackage{bm}
\usepackage{soul}
\usepackage{svg}
\usepackage{algorithm}
\usepackage{algpseudocode}
\usepackage{derivative}
\usepackage{wrapfig}
\usepackage{setspace}
\usepackage[inline]{enumitem}

\usepackage{amsthm}
\newtheorem{theorem}{Theorem}
\newtheorem{remark}{Remark}[theorem]
\newtheorem{lemma}{Lemma}

\usepackage{titlesec}

\newcommand{\para}[1]{\vspace{.05in}\noindent\textbf{#1}}

\begin{document}

\title{Feature Preserving Shrinkage on Bayesian Neural Networks via the R2D2 Prior}

\author{%
  Tsai Hor Chan, Dora Yan Zhang, Guosheng Yin, and Lequan Yu
  \thanks{All authors are affiliated with Department of Statistics and Actuarial Science, University of Hong Kong, Hong Kong. } 
}

\markboth{Journal of \LaTeX\ Class Files,~Vol.~14, No.~8, August~2015}
{Shell \MakeLowercase{\textit{et al.}}: Bare Demo of IEEEtran.cls for Computer Society Journals}
\IEEEtitleabstractindextext{
\begin{abstract}
Bayesian neural networks (BNNs) treat neural network weights as random variables, which aim to provide posterior uncertainty estimates and avoid overfitting by performing inference on the posterior weights.
%
However, selection of appropriate prior distributions remains a challenging task, and BNNs may suffer from catastrophic inflated variance or poor predictive performance when poor choices are made for the priors.
%
%
%
Existing BNN designs apply different priors to weights, while the behaviours of these priors make it difficult to sufficiently shrink noisy signals or they are prone to overshrinking important signals in the weights.
%
To alleviate this problem, we propose a novel R2D2-Net, which imposes the $R^2$-induced Dirichlet Decomposition (R2D2) prior to the BNN weights.
%
%
The R2D2-Net can effectively shrink irrelevant coefficients towards zero, while preventing key features from over-shrinkage. 
%
To approximate the posterior distribution of weights more accurately, we further propose a variational Gibbs inference algorithm that combines the Gibbs updating procedure and gradient-based optimization.
This strategy enhances stability and consistency in estimation when the variational objective involving the shrinkage parameters is non-convex.
%
We also analyze the evidence lower bound (ELBO) and the posterior concentration rates from a theoretical perspective. 
Experiments on both natural and medical image classification and uncertainty estimation tasks demonstrate satisfactory performances of our method.
%
%
\end{abstract}

}

\maketitle

\IEEEdisplaynontitleabstractindextext

\IEEEpeerreviewmaketitle

\section{Introduction}
In the past decades, deep neural networks (DNNs) have shown great success in solving various tasks with high-dimensional features. 
Most of the state-of-the-art (SOTA) DNN architectures adopt frequentist approaches to train a single set of weights.
These models cannot address the epistemic (i.e., model-wise) uncertainties, which may cause overfitting when the number of observations is limited 
\cite{kendall2017uncertainties}.
%
Failure to address the epistemic uncertainties would lead to poor generalization performance for out-of-distribution data, as the model cannot learn robust representations from limited training observations.
Such frequentist methods also 
lack uncertainty estimates as they typically only provide point estimates \cite{kendall2017uncertainties}.
The recent emergence of Bayesian deep learning frameworks provides a practical solution to quantifying uncertainties in 
deep learning models.

Bayesian neural networks (BNNs) refine SOTA deep learning architectures with Bayesian approaches, which enable neural networks to quantify uncertainties arising from models \cite{Jospin2020BNNTutorial, shridhar2019BayesianCNN}. 
BNNs also act as a natural regularization technique that mitigates the bias of the model by performing inference based on posterior distributions of model weights.
%
%
Most of the existing BNN architectures adopt zero-mean multivariate Gaussian distributions as the prior distributions for the weights \cite{shridhar2019BayesianCNN}.
However, such multivariate Gaussian prior distributions often lead to many unnecessary nodes with large variances, which further results in large variances in posterior predictions.
The consequence can be catastrophic because most of the deep BNNs without appropriate priors may underfit the data and thus lead to 
poor predictions \cite{ghosh2019horseshoebnn, tomczak2021collapsedVIBNN}. 
Therefore, variable shrinkage priors (which can shrink coefficients unrelated to tasks to zero) are needed to reduce the noise in coefficients and alleviate the variance inflation issue. 

\begin{figure}
    \centering
    \includegraphics[width=0.35\textwidth]{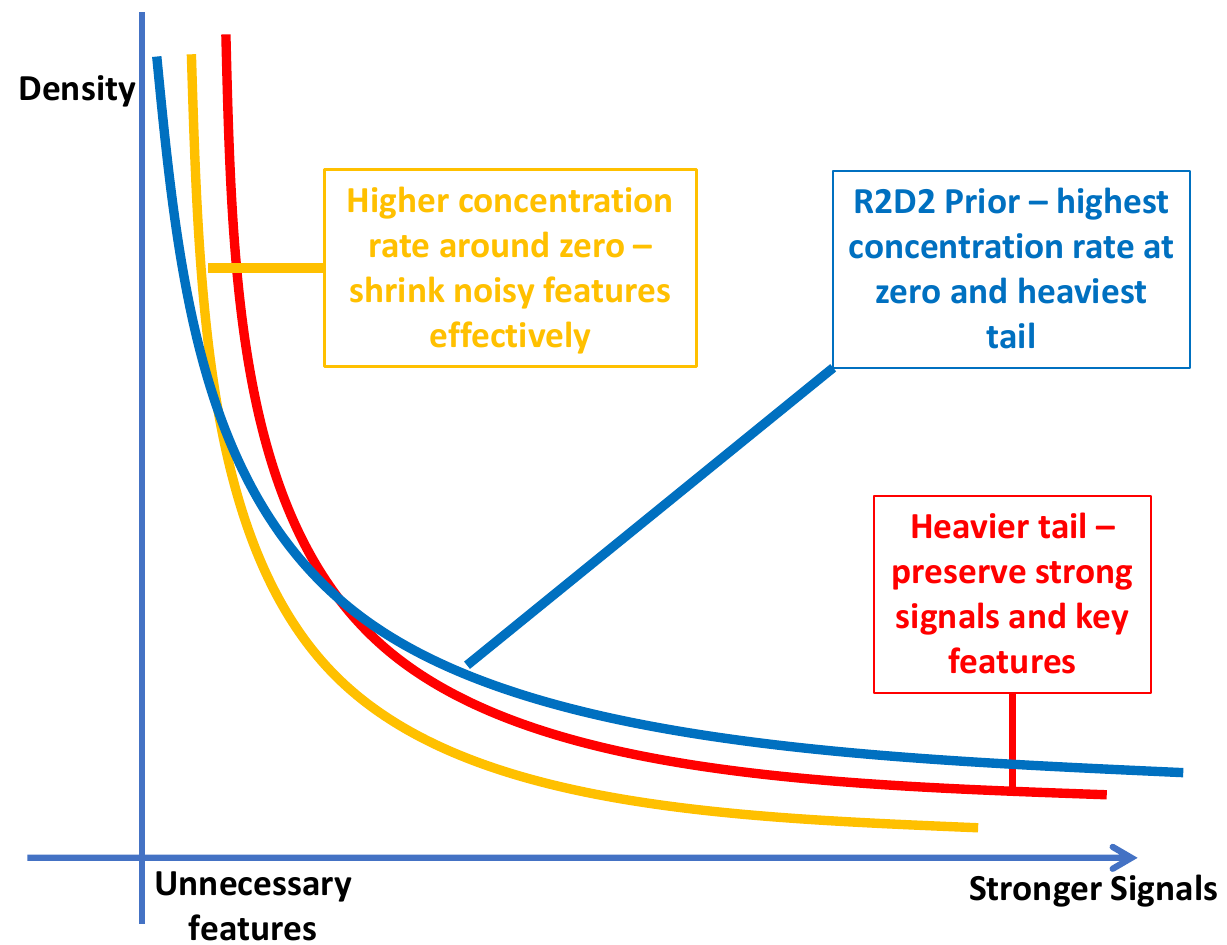}
    \caption{An illustrative comparison of priors with different tail behaviors and concentration rates at zero. 
    A prior with a heavier tail preserves stronger signals by putting more weights on them.
    A prior with a larger concentration rate around zero can shrink unnecessary or trivial features more effectively.
    }
    \label{fig: illustration}
    \vspace{-5mm}
\end{figure}

Recently, several works \cite{ghosh2019horseshoebnn, matsubara2020ridgelet,  popkes2019horseshoeBNNIC, tran2022BNNGPprior} attempt to adopt global--local shrinkage priors to mitigate the problem of large variances. 
These priors are able to shrink the coefficients and alleviate the under-fitting problem in BNNs. 
Although existing shrinkage priors demonstrate superior performance in variable selection, their properties are subject to several limitations. 
For instance, these priors have either a low concentration rate around zero or light tails. 
A low concentration rate around zero leads to weak shrinkage effects, while the variance of prediction remains large. 
Light (or thin) tails under-weigh the effects of large coefficients, which would over-shrink important signals \cite{zhang2020r2d2}. 
In particular, Gaussian distribution has the lightest tail and tends to assign almost zero weight to large signals.
This leads to over-regularization as well as poor feature representation learning, especially when the architecture is deep.
Furthermore, the optimization of shrinkage parameters would be difficult with stochastic variational inference, when the joint distribution of the shrinkage and weight parameters is not convex \cite{ griffin2024structuredShrinkage, hoffman2013SVI}.


The aforementioned limitations motivate us to select a prior that has the highest concentration rate at zero and heaviest tails, which are crucial to predictive models with a large number of parameters --- especially for DNNs. 
Table \ref{tab:gls_priors} presents the concentration rates around zero and the tail thickness of commonly used shrinkage priors.
By comparing the rates of various priors, we are prompted to select the R2D2 prior \citep{zhang2020r2d2} for the neural network weights and propose a novel BNN design --- the R2D2-Net. 
The R2D2-Net is more effective in model selection than designs based on other existing shrinkage priors, because it can choose more important weights in predictive tasks.

\begin{table}[t]
    \centering
    \caption{Tail decay and concentration at zero of commonly used global--local shrinkage priors \cite{zhang2020r2d2}}
    \scalebox{0.8}{
    \begin{tabular}{lcc}
    \toprule 
    Priors & Tail Decay & Concentration at 0 \\ \hline \vspace{2mm}
        Horseshoe & $\mathcal{O}\left(\dfrac{1}{\beta^2}\right)$ &$\mathcal{O}\left(\log\left(\dfrac{1}{|\beta|}\right)\right)$ \\
        Horseshoe+ & $\mathcal{O}\left(\dfrac{\log |\beta|}{\beta^2}\right)$ & $\mathcal{O}\left(\log^2\left(\dfrac{1}{|\beta|}\right)\right)$\\
        Dirichlet--Laplace & $\mathcal{O}\left(\dfrac{|\beta|^{a^*/2 - 3/4}}{\exp\{\sqrt{2|\beta|}\}}\right)$ & $\mathcal{O}\left(\dfrac{1}{|\beta|^{1 - a^*}}\right)$\\
    Generalized Double Pareto & $\mathcal{O}\left(\dfrac{1}{|\beta|^{1 + \alpha}}\right)$ & $\mathcal{O}(1)$\\
        R2D2 & $\mathcal{O}\left(\dfrac{1}{|\beta|^{1 + 2b}}\right)$ & $\mathcal{O}\left(\dfrac{1}{|\beta|^{1 - 2a_\pi}}\right)$\\\bottomrule
    \end{tabular}
    }
    \label{tab:gls_priors}
    {\scriptsize \begin{spacing}{0.5}
Here, $\beta$ represents the model parameters, and $a^*, \alpha, b, \alpha_\pi$ are the respective hyperparameters controlling the shrinkage behaviours, whose precise definitions can be found in the original papers.
\end{spacing}
}
\vspace{-2mm}
\end{table}

\para{Contribution summary:}
(1) We propose a novel BNN design --- the R2D2-Net by specifying an R2D2 prior on the model weights, which improves the shrinkage effect and the predictive performance over existing priors.
(2) 
We develop the stochastic variational Gibbs inference algorithm that
integrates the Gibbs sampling procedure and gradient-based optimization.
It estimates the shrinkage parameters more effectively, particularly when the joint distribution of the shrinkage and weight parameters is not convex. 
%
(3) 
We theoretically analyze the evidence lower bound (ELBO) in the variational inference of BNN and develop analytical forms of the Kullback–Leibler (KL) divergences of the shrinkage parameters.
Theoretical analysis validates that the R2D2-Net possesses the minimax posterior concentration rate under the polynomial boundedness assumption.
(4) Extensive synthetic and real data experiments demonstrate the outstanding performance of R2D2-Net on inference, predictive and uncertainty estimation tasks compared with various combinations of existing priors and inference algorithms.
Codes are available at \url{https://github.com/HKU-MedAI/r2d2bnn}.
%
\section{Related Works}

\para{Global--Local Shrinkage Priors.}
High-dimensional regression often suffers from the curse of dimensionality, which 
motivates novel approaches to  shrinkage of coefficients and variable selection.
Global--local shrinkage priors are a class of shrinkage priors that can be essentially expressed as a global--local scale Gaussian family.
Existing shrinkage priors possess desirable theoretical and empirical properties that can effectively perform coefficient shrinkage. 
\citeauthor{carvalho2009horseshoe} proposed the Horseshoe prior, which exhibits Cauchy-like flat and heavy tails and maintains a high concentration rate at zero. 
Although the Horseshoe prior and its variants \cite{bhadra2017horseshoe+, piironen2017regularizedhorseshoe} are shown to be able to shrink the coefficients, their tail thickness and concentration rates at zero are less desirable compared with some recently proposed global--local shrinkage priors. 
A higher concentration rate at zero allows the model to shrink unnecessary coefficients toward zero more aggressively, and a heavier tail can avoid shrinking key coefficients that have large values with strong signals. 
\citeauthor{zhang2020r2d2}  proposed the $R^2$-induced Dirichlet Decomposition (R2D2) prior, which specifies a prior based on the $R^2$ value of a model fit.
The R2D2 prior demonstrates optimal behaviors both in the tails and concentration at zero, which potentially provides the best shrinkage performance while preserving the important signals in the weights.


\para{Bayesian Neural Networks.}
BNNs specify prior distributions on the weights and bias parameters of the neural network. A vanilla BNN assumes a zero-mean multivariate Gaussian distribution on the parameters. 
The MC Dropout approach \cite{gal2016mcdropout} randomly drops out weights to produce posterior samples from a trained frequentist neural network.
%
%
Moreover, the variance inflation exacerbates as the number of layers increases, making it extremely difficult to build and optimize deep BNNs \cite{dusenberry2020rank1BNN}. 
Most of the existing works focus on small architectures (e.g., LeNet) and small datasets (e.g., CIFAR 100), while several works \cite{dusenberry2020rank1BNN, lakshminarayanan2017deepensembles} attempt to scale up to more modern and deeper architectures (e.g., ResNet101).

To address the above issues, sparsification methods have been adopted to shrink unnecessary neurons to prevent variance inflation.
Utilization of sparsity--induced priors \cite{louizos2017BCDL} has become a more popular approach than variational dropout methods \cite{molchanov2017sparseVD, smith2018ensembleMC}.
\citeauthor{ghosh2019horseshoebnn} proposed to place the Horseshoe prior on the variances of weights to resolve the large prediction variance problem. 
However, due to the relatively low concentration rate around zero, the shrinkage effect is compromised. 
%
Moreover, the relative lighter tails of the Horseshoe prior than R2D2 limits its capability to preserve important signals, which likely leads to over-shrinkage. 
%
%

\para{Posterior Computation of BNNs.}
Conventional methods for computing BNN posteriors are mostly sampling-based \cite{chen2014HMC,  nemeth2021SGMCMC, welling2011SGLD, yu2023SGSampling}, which are practical for neural networks since they follow closely the stochastic optimization schema \cite{yu2023SGSampling}. 
Existing works on sampling-based approaches can be categorized into two prominent families: Langevin dynamics \cite{welling2011BNNLangevin} and Hamiltonian dynamics \cite{chen2014HMC, girolami2011riemannHMC}. 
Despite their success, these sampling-based methods are often considered inefficient in terms of computational complexity.
When tackling large-scale problems, simplified computational metrics need to be introduced for better computational efficiency \cite{yu2023SSGRLD}.
Recently, there have been approaches utilizing variational inference (VI)  to learn the probabilistic assumptions in BNNs.

Classical BNN training paradigms widely adopt a mean field VI approach to approximating the posteriors, which assume independent marginal distributions \cite{farquhar2020radialBNN, garner2020bayesianrnn, ghosh2019horseshoebnn, 
molchanov2017sparseVD,
rudner2022FSVI,  
shridhar2019BayesianCNN}.
The VI typically approximates a posterior distribution $p(\bm \theta | \bm y)$ by a variational posterior distribution $q$ obtained from a candidate set $\mathcal{Q}$ by maximizing an evidence lower bound (ELBO):
\begin{equation} \label{eq: ELBO}
    \max_{q \in \mathcal{Q}} \mathbb{E}_{\bm \theta \sim q}[\log  p(\bm y|\bm \theta)] - \text{KL} (q \Vert \pi),
\end{equation}
where $
\text{KL}(q \Vert \pi) = \mathbb{E}_{q \in \mathcal{Q}}[\log p(\bm \theta| \cdot)] + \mathbb{H}[\pi(\bm \theta)],
$  is the KL divergence between $q$ and the prior distribution $\pi$, 
$\mathbb{H}(\pi(\bm \theta))$ is the entropy of the distribution, and $p(\bm y|\bm \theta)$ is the likelihood.
%
Most of the works \cite{farquhar2020radialBNN,
gal2016mcdropout,
ghosh2019horseshoebnn, shridhar2019BayesianCNN} assume Gaussian prior distributions on weights and hence the KL divergence can be approximated by
$
    \text{KL}(q \Vert \pi) = \sum_{j, l} \text{KL}(q(w_{jl} | \cdot) \Vert \pi(w_{jl}|\cdot)),
$
\begin{figure*}
    \centering
    \includegraphics[width=0.8\textwidth]{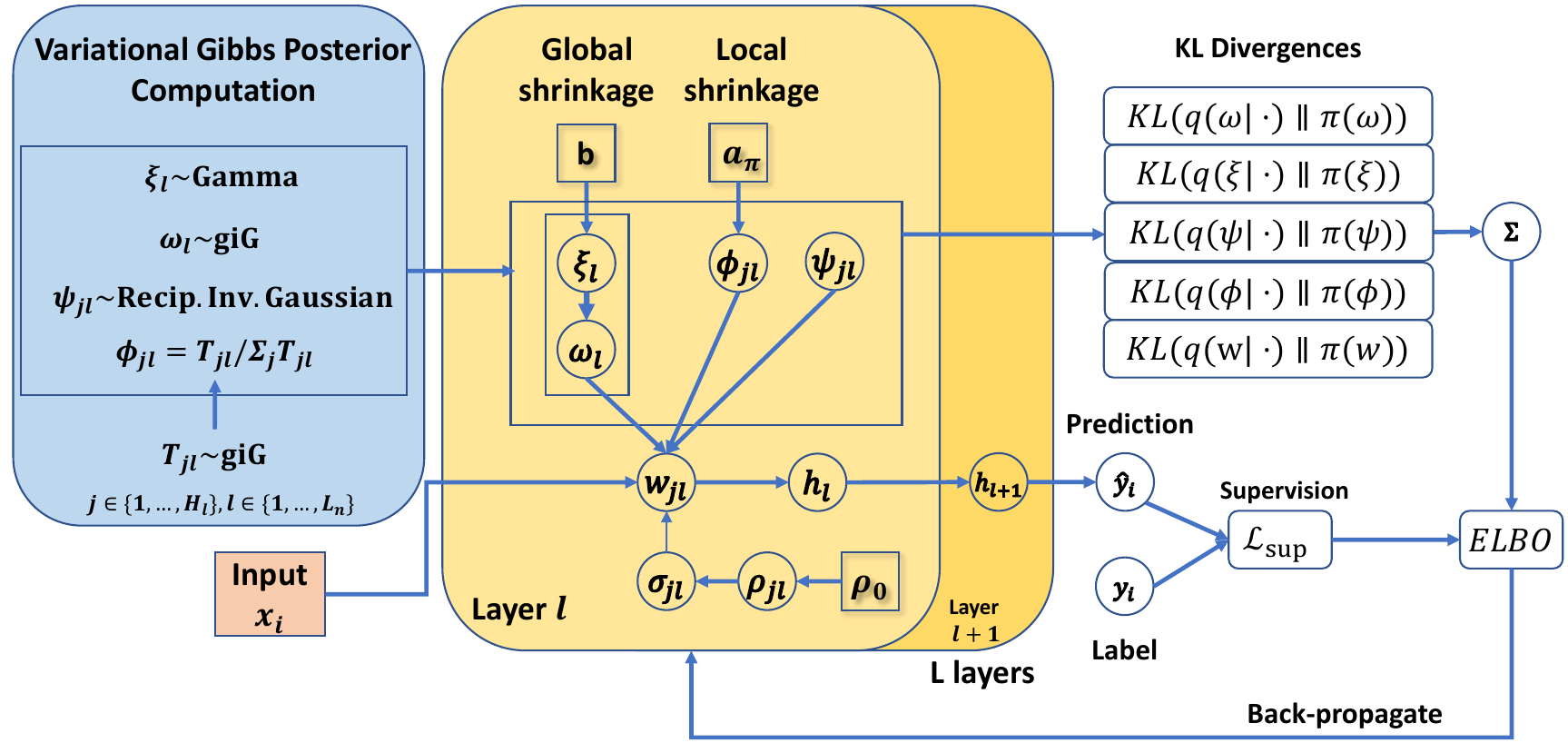}
    \caption{
    Overview of the proposed R2D2-Net with the yellow part representing the graphical model of each neuron and the blue part summarizing the variational Gibbs inference for computing the posterior distribution of weights.
    }
    \label{fig:r2d2_graphical_model}
    \vspace{-3mm}
\end{figure*}
where $q(w_{jl} | \cdot)$ is the variational posterior and $\pi(w_{jl} | \cdot)=\mathcal{N}(w_{jl}|0, 1)$ is the standard normal prior distribution on weights.
However, Gaussian assumption is made for convenience and the estimation of KL suffers from large approximation error. 
By comparing the KL divergence of the analytical distributions of the hierarchical prior (e.g., Horseshoe, R2D2), a more accurate approximation of the ELBO can be obtained.

\para{Sparsifying Neural Networks.}
Another related field of our work is neural sparsification, which focuses on compressing neural networks to prune unnecessary neurons and improve the space efficiency \cite{han2015DC, louizos2017BCDL, molchanov2017sparseVD,  srinivas2016GD}.
Sparsity-induced prior is also a popular choice in this field \cite{ghosh2019horseshoebnn, louizos2017BCDL}.
%
%
%
%
%
%
Despite similarities in these approaches, our work focuses on a BNN design with shrinkage priors which can improve its capability of predictive and uncertainty estimation instead of compressing the existing architectures.

\section{Preliminaries}
\textbf{Deep Neural Network (DNN).} A DNN with $L$ layers can be written as
\begin{align*}
    f_l(\bm x) = \dfrac{1}{\sqrt{d_{l-1}}} W_l \phi (f_{l-1}(\bm x))  + b_l, \quad l \in \{ 1, \ldots , L\},
\end{align*}
where $\bm x$ is the input feature vector, $\phi$ is a nonlinearity activation function (e.g., the rectified linear unit (ReLU), $\phi(a) = \max (0, a)$), $d_{l-1}$ is the dimension of the input of layer $l-1$, $b_l \in \mathbb{R}^{d_l}$ is a vector containing the bias parameters for layer $l$, and $W_l$ is the weight tensor. 
For linear layers, we have $W_l \in \mathbb{R}^{d_l \times d_{l-1}}$,
and for convolutional layers, $W_l \in \mathbb{R}^{d_l \times d_{l-1}\times k_0 \times k_0} $, where $k_0$ is the kernel size.
Let $\mathbf{w}_l = \{W_l, b_l\}$ denote the set of weight and bias parameters of layer $l$, and let $w_{jl}$ denote the $j$-th element of the parameter vector at layer $l$, and let $p_l = |\mathbf{w}_l|$, the size of set $\mathbf{w}_l$. 
The trainable network parameters are summarized as $\bm \theta = \{\mathbf{w}_l\}_{l=1}^L$.

\para{Notation for Sparsity-induced BNN.}
Let $K_n=|\bm \theta|$ denote the total number of parameters, and let $D_n$ denote the dimension of the input and they are assumed to grow with the training size $n$. We further allow $L_n$, the number of layers, increasing with $n$. Let $H_l$ be the number of hidden units at layer $l$ with $\bar{H} = \max_{1\leq l \leq L_n-1} H_l$. Let $\bm \gamma = \{1_{w_{jl} > 0 }: j \in \{1, \ldots, H_l\}, l \in \{1, \ldots, L_n\} \}$ denote the connectivity of each parameter $w_{jl}$.

\para{Bayesian Neural Network (BNN).} A BNN specifies a prior $\pi(\bm \theta)$ on the trainable weights $\bm \theta$. 
Given the dataset $\mathcal{D} = \left\{ \bm x_i, y_i\right\}_{i=1}^n$ of $n$ pairs of observations, 
we aim to estimate the posterior distribution of the weights,
\begin{align*}
    p(\bm \theta | \mathcal{D}) = \dfrac{\pi (\bm \theta) \prod_{i=1}^n p(y_i|f (\bm \theta, \bm x_i))}{p(\mathcal{D})},
\end{align*}
where $p(y_i|f (\bm \theta, \bm x_i))$ is the likelihood function and $p(\mathcal{D})$ is the normalization term.
%

\para{Multivariate Gaussian.}
The density of a $p$-dimensional multivariate Gaussian distribution is defined as
\begin{align*}
       & f(\bm x; \bm \mu, \bm \Sigma)\\ =
       &\dfrac{1}{(2 \pi)^{p/2}|\bm \Sigma|^{1/2}} \exp\Big\{-\dfrac{1}{2} (\bm x - \bm \mu)^\top \bm \Sigma^{-1}(\bm x - \bm \mu)\Big\},
\end{align*}
where $\bm \mu \in \mathbb{R}^p$ is the mean vector and $\bm \Sigma \in \mathbb{R}^{p \times p}$ is the covariance matrix. 

\para{Generalized Inverse Gaussian.}
The generalized inverse Gaussian distribution is denoted as $Z \sim \text{giG} (\chi, \rho, \lambda_0)$, which has the density function,
\begin{align*}
    f(z;\chi, \rho, \lambda_0) = \frac{(\rho / \chi)^{\lambda_0 / 2}}{2K_{\lambda_0}(\sqrt{\rho \chi})} z^{\lambda_0 - 1} \exp \{ -(\rho z + \chi / z) / 2\},
\end{align*}
where $K_{\lambda_0}(\cdot)$ is a modified Bessel function of the second kind. Specifically, an inverse Gaussian distribution of the form 
$
    f(x;\mu, \lambda) = \left( \frac{\lambda}{2 \pi x^3} \right)^{1/2} \exp \left( \frac{-\lambda(x-\mu)^2}{2\mu^2x} \right)
$
is a special case of giG with $\rho = \lambda / \mu^2$, $\chi = \lambda$, and $\lambda_0 = - {1}/{2}$. 

\vspace{-2mm}
\section{Methodology}
To achieve the best variable shrinkage performance, we impose the R2D2 prior on the neural network weights, leading to the R2D2-Net.
\begin{table*}[h]
    \centering
    \caption{Analytical forms of KL-divergences of the shrinkage parameters ($\xi_l$, $\omega_l$,  $\psi_{jl}$)}
    \scalebox{0.85}{
       \begin{tabular}{cllc}
    \toprule 
    & \textbf{Prior }$\pi$ & \textbf{Variational Posterior} $q$ & \textbf{Closed Form of KL-divergence}\\  \hline
    $\xi_l$ & Gamma & Gamma & $\mathbb{E}_{q}\left[\log \left(\dfrac{(1 + \omega_l)^{a_l + b_l}}{\Gamma(a_l + b_l)}\xi_l^{a_l + b_l - 1}e^{-(1 + \omega_l) \xi_l}\right)\right] - \mathbb{E}_{q}\left[\log \left(\dfrac{1}{\Gamma(b_l)}\xi_l^{b_l - 1}e^{- \xi_l}\right)\right]$  \\
    $\omega_l$& Gamma & Generalized InvGaussian &  $\mathbb{E}_{q}\left[\log \left(\dfrac{(\rho/\chi)^{\lambda_0 /2}}{2K_{\lambda_0}(\sqrt{\rho\chi})}\omega_l^{\lambda_0-1}e^{(-\rho\omega_l+\chi/\omega_l)/2}\right)\right]
     - \mathbb{E}_{q}\left[\log \left(\dfrac{\xi_l^{a_l}}{\Gamma(a_l)}\omega_l^{a_l - 1}e^{-\omega_l\xi_l}\right)\right]$\\
    $\psi_{jl}$ & Exp & Reciprocal InvGaussian & $\mathbb{E}_{q}\left[\log \left(\dfrac{1}{\psi_{jl}\sqrt{2\pi}}\exp\left(\dfrac{(1 - \psi_{jl}\mu)^2}{2\psi_{jl}\mu}\right)\right)\right]
     - \mathbb{E}_{q}\left[\log \left(\dfrac{1}{2}e^{-\frac{1}{2}\psi_{jl}}\right)\right]$\\
    \bottomrule
    \end{tabular}}
    \label{tab:kl-divregences}
    \vspace{-2mm}
\end{table*}
By placing the R2D2 prior on the weights, irrelevant weights can be shrunk effectively towards zero and important weights can be well preserved. 
We also propose a variational Gibbs inference procedure and develop analytical forms of KL divergences of the shrinkage parameters to obtain better estimates of posterior distributions of the weights.
\vspace{-3mm}
\subsection{The R2D2-Net} \label{subsec: r2d2net}

Consider a linear model,
\begin{align} \label{eq: linear model}
    y_i = \bm x_i^\top \bm \beta + \epsilon_i, \ \  i=1,\ldots, n, 
\end{align}
where $y_i$ is the response, $\bm x_i$ is the $p$-dimensional vector of covariates for the $i$-th observation,  
$\bm \beta = (\beta_1, \ldots, \beta_p)^\top$ is a vector of coefficients, and $\epsilon_i$ is the error term. 
The R2D2-Net specifies a prior on the $R^2$ from the model fit of (\ref{eq: linear model}). The $R^2$ of linear prediction is given by
\begin{align*}
    R^2(\bm \beta) = \dfrac{\text{var}(\bm X^\top \bm \beta)}{\text{var}(\bm X^\top \bm \beta) + \sigma^2},
\end{align*}
where $\bm \beta$ can be viewed as the weight tensor of the convolutional or the linear layer and $\bm X \in \mathbb{R}^{p \times n}$ is the data matrix. 
 By specifying a beta prior on $R^2(\bm \beta)$, the marginal R2D2 prior has the form,
\begin{equation} \label{eq: r2d2_prior}
\begin{split}
     \beta_j | \psi_j, \omega, \phi_j \sim \mathcal{N}(0, \psi_j\phi_j\omega\sigma^2/2), \  \psi_j \sim {\rm Exp}(1/2), \\ 
    \bm \phi  \sim \text{Dir}(a_\pi, \ldots, a_\pi), \  
    \omega | \xi \sim \text{Ga}(a, \xi), \  \xi \sim \text{Ga}(b, 1),
    \end{split}
\end{equation}
where Exp denotes an exponential distribution, Ga a Gamma distribution, and Dir a Dirichlet distribution, and $\phi_j$ is the $j$-th element of $\bm \phi$.
The R2D2 prior has the highest concentration rate at zero and heavier tails than other global--local priors \cite{zhang2020r2d2}. 
Therefore, it can more effectively shrink the coefficients of unimportant covariates 
to zero. 
For coefficients that have large signals (i.e., large norms), the heavy-tail nature of the R2D2 prior is able to avoid over-shrinking these coefficients, thus preserving the ability to extract key features from the input data. 

With the marginal distributions of weights in Eq. (\ref{eq: r2d2_prior}), we can formulate the layers of the R2D2-Net.
%
We assign the R2D2 distribution in Eq. (\ref{eq: r2d2_prior}) as the prior for $w_{jl}$, the $j$-th element of parameters of the $l$-th layer $\mathbf{w}_l$.
We can then compose the R2D2-Net by specifying a combination of convolutional layers and linear layers.
\vspace{-4mm}
\subsection{Variational Gibbs Inference for Optimization}
We adopt a mean field approach to computing the ELBO by factorizing $q (\bm \theta)$ into a product of the marginal distributions of all neurons.
First, we update $\mathbf{w} $ and $\bm \rho$ by back-propagating the ELBO in Eq. (\ref{eq: ELBO}). 
We initialize the weight parameters $\mathbf{w}_l$ with a reparameterized Gaussian distribution,
$
    w_{jl} \sim \mathcal{N}(\mu_{jl}, \sigma_{jl}^2\psi_{jl}\phi_{jl} \omega_l),
$
where each standard deviation $\sigma_{jl}$ is reparameterized by introducing a parameter $\rho_{jl}$ such that $\sigma_{jl} = \log \left(1 + e^{\rho_{jl}}\right)$.
We assign an individual variance term $\sigma_{jl}$ to each weight, which is different from \citeauthor{zhang2020r2d2} \cite{zhang2020r2d2} who assume the same $\bm \sigma_l = \sigma_l \bm 1$ for all weight parameters in layer $l$.
The distribution of $\sigma_l$ in \citeauthor{zhang2020r2d2} is updated by the regression MSE, which is analogous to learning the variance of neurons by back-propagation of task-specific loss.
Therefore, under the deep learning setting, we distinctively specify a variance parameter $\bm \sigma_l$
for each neuron and learn them by back-propagating the task-specific loss.
%
We set the prior values of $\bm \psi=\{\psi_{jl}\}_{j=1}^{p_l}{}_{l=1}^L, \bm \phi = \{\phi_{jl}\}_{j=1}^{p_l}{}_{l=1}^L, \bm \omega = \{\omega_{l}\}_{l=1}^L, \bm \xi = \{\xi_{l}\}_{l=1}^L$ with the prior distribution defined in Eq. (\ref{eq: r2d2_prior}) and $\mu_{jl} = 0, \rho_{jl} = \rho_0 $ for the first step.
With the weight parameter samples, we are able to compute the ELBO using Eq. (\ref{eq: ELBO}).
The trainable parameters $\mathbf{w} $ and $\bm \rho$ can be updated by back-propagating the ELBO.

We then update shrinkage parameters using the updated $\mathbf{w} $ and $\bm \sigma$.
Following the Gibbs sampling procedures proposed by \citeauthor{zhang2020r2d2}, we develop our variational Gibbs inference algorithm to update the shrinkage parameters alternatively using their individual posterior distributions.
We first sample $\psi_{jl}, \omega_l $ and $\xi_l$,
 \begin{align*}
         \psi^{-1}_{jl} & \mid  w_{jl}, \phi_{jl}, \sigma_{jl}^2 \ \ \  \\
         & \sim \text{InvGaussian} \big(\mu=\sqrt{\sigma_{jl}^2 \phi_{jl} \omega_l / 2}/|w_{jl}|, \lambda=1\big),
\end{align*}
\begin{align*}
        \omega_l & \mid \mathbf{w}_l, \bm \phi_{l}, \xi_l, \bm \sigma_l^2 \\
        & \sim  \text{giG}\bigg(\chi = \sum_{j=1}^{p_l} 2 w_{jl}^2/(\sigma_{jl}^2 \phi_{jl} \psi_{jl}), 
        & \rho = 2\xi_l, \lambda_0 = a_l - \frac{p_l}{2}\bigg),
\end{align*}
\begin{align*}
        \xi_l \mid \omega_l & \sim \text{Ga}(a_l + b_l, 1 + \omega_l) .\\
\end{align*}
To sample $\bm \phi_{l} \mid \mathbf{w}_l, \bm \psi_l, \xi_l, \bm \sigma_l^2,$ we first draw $T_{1l}, \ldots, T_{p_l l}$ independently with $T_{jl} \sim \text{giG} (\chi = 2 w_{jl}^2/(\sigma_{jl}^2 \psi_{jl}), \rho = 2\xi_l, \lambda_0 = a_l - \frac{p_l}{2}) $, and then set $\phi_{jl}= \frac{T_{jl}}{T_l}$ with $T_l = \sum_j T_{jl}$.
We repeat the above steps to train the R2D2-Net iteratively till convergence or early stopping criteria are met (e.g., the loss does not decrease).
Algorithm \ref{alg: var-gibbs} presents the detailed workflow of the variational Gibbs inference, which leverages the advantages of both posterior computation and gradient-based estimation to obtain better approximation of the shrinkage parameters. 

\begin{algorithm*}[t]
\begin{algorithmic}[1]
\Statex \textbf{Input:} 
\Statex Number of layers $L$; Prior distributions of weight parameters $\pi(\bm \theta)$; 
\Statex Total numbers of parameters of each layer $\{p_l\}_{l=1}^L$;
\Statex Local shrinkage parameters $\bm \phi = \{\phi_{jl}\}_{j=1}^{p_l}{}_{l=1}^L , \bm \psi = \{\psi_{jl}\}_{j=1}^{p_l}{}_{l=1}^L$;  
\Statex  Global shrinkage parameters $\bm \omega = \{\omega_{l}\}_{l=1}^L, \bm \xi = \{\xi_{l}\}_{l=1}^L$;
\Statex Input data $\mathcal{D} = \left\{\bm x_i, y_i\right\}_{i=1}^n$;
\Statex Hyperparameters $a_\pi, \rho_0,$ $b$.
\Statex \textbf{Output:} 
\Statex The posterior distribution of the weights $p(\bm \theta | \mathcal{D})$.
\State Initialize $\pi (w_{jl}) \sim \mathcal{N}( 0, (\log \left(1 + e^{\rho_{0}}\right))^2), \ a_l = p_l a_\pi, \ b_l = b, \ \mu_{w_{jl}} = 0$
\State $\bm h_1 = \bm x_i$
\State Sample $\psi_{jl} \sim \text{Exp}({1}/{2}),\   \phi_{jl} \sim \text{Dir}(a_\pi, \ldots, a_\pi),\  \xi_l \sim \text{Ga}(b_l, 1), \ \omega_l | \xi_l \sim \text{Ga}(a_l, \xi_l)$
\For{each step}
\For{$l$ in $1:L$}
\For{$w_{jl}$ in $\mathbf{w}_l$}
    \State Sample  $w_{jl} \sim \mathcal{N}(\mu_{w_{jl}}, \psi_{jl}\phi_{jl}\omega_l\sigma_{jl}^2/2)$  \Comment{Sample weights}
    \State Sample $\rho_{jl} \sim \mathcal{N}(\mu_{\rho_{jl}}, \sigma^2_{\rho_{jl}})$ 
    \State Set $\sigma_{jl} = \log \left(1 + e^{\rho_{jl}}\right)$ \Comment{Reparameterized Gaussian}
    \State Sample $\omega\sim\text{giG}(\chi = \sum_{j=1}^{p_l} 2 w_{jl}^2/(\sigma^2 \phi_{jl} \psi_{jl}), \rho = 2\xi_l, \lambda_0 = a_l - \frac{p_l}{2})$
    \State Sample $\xi_l \sim \text{Ga}(a_l + b_l, 1 + \omega_l)$
    \State Sample $\psi_{jl}^{-1} \sim $ InvGaussian$(\mu=\sqrt{\sigma_{jl}^2 \phi_{jl} \omega_l / 2}/|w_{jl}|, \lambda=1)$
    \State Sample  $T_{jl} \sim \text{giG} (\chi = 2 w_{jl}^2/(\sigma_{jl}^2 \psi_{jl}), \rho = 2\xi_l, \lambda_0 = a_l - \frac{p_l}{2})$
    \State Set $\phi_{jl} = T_{jl} / \sum_j T_{jl}$
\EndFor
    \State Compute $\bm h_{l+1} = \mathbf{w}_{l} \bm h_{l} + \text{bias}_l$
\EndFor
    \State Obtain prediction $\bm \hat{y}_n$ from $\bm h_{L-1}$
    \State Compute the supervision loss,  $\text{KL}(q \Vert \pi)$ (Table \ref{tab:kl-divregences}), and the ELBO.
    \State Back-propagate the ELBO to update the mean and variance of $\bm \theta$.
\EndFor
\State \Return Posterior distribution $p(\bm \theta | \mathcal{D})$.
\end{algorithmic}
\caption{The Variational Gibbs Inference Algorithm.}
\label{alg: var-gibbs}
\end{algorithm*}

\vspace{-3mm}
\subsection{Estimation of KL Divergences with Variational Posterior Distributions}

In light of the importance of obtaining an accurate estimate of the KL loss in variational inference, we utilize the full posterior distribution obtained in variational Gibbs inference and the R2D2 prior to obtain a more accurate estimate of the KL loss.
The KL divergence of the variational posterior $q$ and the prior $\pi$ can be divided into several components as follows,
\begin{flalign*}
    \text{KL}(q(\bm \theta | &\cdot)\Vert\pi(\bm \theta)) 
    = \text{KL}(q(\bm \xi | \cdot) \Vert \pi(\bm \xi))\\ 
    &+ 
    \text{KL}(q(\bm \omega | \cdot) \Vert \pi(\bm \omega)) 
    + \text{KL}(q(\bm \psi |\cdot) \Vert \pi(\bm \psi)) \\
    &+ \text{KL}(q(\bm \phi |\cdot) \Vert \pi(\bm \phi))  +
    \text{KL}(q(\mathbf{w} |\cdot) \Vert \pi(\mathbf{w})).
\end{flalign*}
We can obtain the closed-form solutions of the KL divergences for $\bm \xi, \bm \omega,$ and $\bm \psi$. 
We approximate the KL divergence of $\bm \phi$ using samples from the variational posterior distribution $q(\bm \phi|\cdot)$. 
Table \ref{tab:kl-divregences} presents the closed forms of the KL divergences on the shrinkage parameters $\xi_l$, $\omega_l$, and  $\psi_{jl}$.
The closed forms in Table \ref{tab:kl-divregences} can be obtained by using  $\mathbb{E}(X), \mathbb{E}(X^{-1})$ and $\mathbb{E}(\log X)$ for $X \sim \text{giG}$, which are given in the supplementary materials together with the detailed derivations of the KL losses.

\section{Posterior Consistency of R2D2-Net }
Posterior consistency measures the speed of posterior concentration rates around its true density function, which is a major metric for validating a prior. 
We extend the theoretical results in \citeauthor{zhang2020r2d2} \cite{zhang2020r2d2}, which establish theoretical properties of the R2D2 prior, and the results from \citeauthor{sun2022spikeandslab} \cite{sun2022spikeandslab}, which formulate the convergence rates of sparse DNNs under general priors.

\para{Regularity Conditions.}
Let $\mu (\bm \theta, \bm \gamma, \bm x)$ be the prediction of a BNN given an input $\bm x$, where $\bm \theta$ is the weights and biases of the BNN and $\bm \gamma$ is the set of all shrinkage parameters of the BNN.
The regularity conditions of sparse DNNs are specified as follows.

\begin{enumerate}[label=A.\arabic*]
\item The input $\bm x$ is bounded by 1 entry-wisely, i.e., $\bm x \in \Omega = [-1, 1]^{D_n}$, and the density of $\bm x$ is bounded in its support $\Omega$ uniformly with respect to $n$.

\item The unknown regression mean function $\mu^*(\bm x)$ can be well approximated by a sparse DNN model such that $\mu(\bm \theta^*, \bm \gamma^*, \bm x)$ satisfies the following conditions:

\begin{enumerate}[label=A.2.\arabic*]
    \item $\Vert \mu^*(\bm x) - \mu(\bm \theta^*, \bm x) \Vert_{L^2(\Omega)} \leq \varpi_n$ where the approximation error $\varpi_n \to 0$ as sample size $n \to \infty$.
    \item The network structure satisfies: $r_n L_n \log n + r_n \log \bar{H} + s_n \log D_n \leq C_0 n^{1 - \epsilon}$, where $0 < \epsilon <1$ is a small constant, $r_n = |\bm \gamma^*|$ denotes the connectivity of $\bm \gamma^*$, and $s_n$ denotes the input dimension of $\bm \gamma^*$.
 \item The network weights are polynomially bounded:  $\Vert \bm \theta^*\Vert_\infty < E_n$, where $E_n = n^{C_1}$ for some constant $C_1 > 0$.
\end{enumerate}
\item The activation function $\phi$ is Lipschitz continuous with a Lipschitz constant of 1. 
\end{enumerate}

To ensure the minimax rate, we further assume the polynomial bounding condition from \cite{bolcskei2019polynomialboundDNN,sun2022sparseDNNTheories}, which is slightly stronger than similar conditions in the existing literature on sparsity-induced regression problems. 
\citeauthor{bolcskei2019polynomialboundDNN} \cite{bolcskei2019polynomialboundDNN} proved that if the network parameters are bounded in absolute value by some polynomial $g(r_n)$, i.e., $\Vert \bm \theta^*\Vert_\infty \leq g(r_n)$, then the approximation error $\varpi = \mathcal{O}(r_n^{-\alpha^*})$ for some constant $\alpha^*$.
Condition A.1 is a typical assumption for posterior consistency (e.g., see \cite{  polson2018posteriorconcentr,sun2022spikeandslab,zhang2020r2d2}) where all bounded data can be normalized to satisfy this assumption. Condition A.3 is satisfied by many conventional activation functions such as sigmoid, tanh, and ReLU.

\begin{theorem} \label{theorem: 1}
Consider a DNN with $L_n$ layers and at most $K_n$ connections, where both $L_n$ and $K_n$ are increasing with $n$.
    Let $k_n \asymp \sqrt{r_n(\log K_n) / n}$ and denote $\mathbb{P}^*$ and $\mathbb{E}^*$ the respective probability measure and expectation with respect to the data $\mathcal{D}$.
        Assume that conditions A.1--A.3 hold,
        if the hyperparameter $a_{\pi} \leq \log (1 - D_n^{-1 + u}) / (2 \log k_n)$, and $E_n / ( L_n \log n + \log \bar{H} )^{1/2} \lesssim b \lesssim n^\alpha$ for some constant $\alpha > 0$, then there exists an error sequence $\epsilon_n^2 = \mathcal{O}(\varpi_n^2) + \mathcal{O}(\zeta_n^2)$, with $\zeta_n^2 = [r_n L_n \log n + r_n \log \bar{H} + s_n \log D_n] / n$, such that $\lim_{n \to \infty} \epsilon_n = 0$ and $\lim_{n \to \infty} n \epsilon^2_n = \infty$,  and the posterior distribution satisfies
    \begin{enumerate}
        \item 
        $\mathbb{P}^*\big\{ p[d(p_{\bm \theta}, p_{\mu^*}) > 4 \epsilon_n | \mathcal{D}] \geq 2 e^{-cn \epsilon_n^2} \big\} \leq 2 e^{-cn \epsilon_n^2}$,
        \item 
$\mathbb{E}^*_{\mathcal{D}} \left\{ p[d(p_{\bm \theta}, p_{\mu^*}) > 4 \epsilon_n | \mathcal{D}]\right\} \leq 4 e^{-2cn \epsilon_n^2}$,
\end{enumerate}
for sufficiently large $n$, where $c$ is a constant, $d$ is the Hellinger distance between two density distributions, $p[\cdot]$ represents the posterior distribution, $p_{\mu^*}$ denotes the underlying true data distribution, and $p_{\bm \theta}$ denotes the data distribution reconstructed by the BNN based on its posterior samples.
\end{theorem}

Theorem \ref{theorem: 1} establishes the Bayesian contraction rate for the R2D2-Net under the Hellinger metric.
The detailed proof of this theorem follows \cite{sun2022consistentsparse, sun2022spikeandslab} and is provided in the supplementary materials. 

\begin{remark}
The minimax $\epsilon$-convergence  rate may not hold if we do not assume $E_n$ to be bounded by a polynomial.
When the assumption is loosen to $\log (E_n) = \mathcal{O}(\log D_n)$, 
we have the $\epsilon$-neighbouhood contracting rate $\epsilon_n = (a_n \log D_n)^{1/2}/ n $ for the Horseshoe prior \cite{armagan2013posteriorRatesShrinkages} and $\epsilon_n = (a_n \log D_n)^{1/2}/ \sqrt{n} $ for the spike-and-slab prior \cite{sun2022spikeandslab} under the linear regression settings.
If the polynomial assumption is violated, the R2D2 prior also provides a near-minimax $\epsilon$-convergence rate under a relaxed assumption given by \citeauthor{zhang2020r2d2} \cite{zhang2020r2d2} with tighter assumptions on $\alpha_\pi$ and $b$, where the spike-and-slab prior shares the same rate under such assumptions \cite{gan2022gr2d2, sun2022spikeandslab}.
\end{remark}

\section{Simulation Study} 
\label{sec: simulations}
We first apply our method to simulated scenarios to validate the predictive and shrinkage performance of the R2D2-Net.
We control the depth to observe how the performance varies as the depth of the network increases.
We also test the inference performance using Hamiltonian Monte Carlo (HMC) fitted weights on the respective prior, which is treated as the oracle truth.

\vspace{-2mm}
\subsection{Experimental Setup}
\vspace{-3mm}
\para{Scenarios.}
We generate the data  $\mathcal{D} = \left\{\bm x_i, y_i\right\}_{i=1}^n$ with 
$n=10000$ and each data point $x_{ij}$, the $j$-th component of $\bm x_i$, is sampled from a uniform distribution $\mathcal{U}(-5, 5)$, and the noise $\epsilon_i \sim \mathcal{N}(0, 3^2)$. We design three simulation scenarios:
(1) \textbf{Polynomial case}: $y_i = x_{i}^3 + \epsilon_i$;
(2) \textbf{Low-dimensional non-linear regression}: $y_i = x_{i1}x_{i2} + x_{i3}x_{i4} + \epsilon_i$;
(3) \textbf{High-dimensional non-linear regression}: $y_i = f(\bm x_i) + \epsilon_i$, where $f$ is a two-layer perceptron with randomly initialized weights and \textit{Relu} nonlinearity.
Additional scenarios and results are presented in the supplementary materials. 
In contrast to other scenarios, the data in Scenario 3 are generated from a randomly initialized neural network. 
The features are hence mostly noise (or trivial) features and shrinkage methods are expected to underperform as they shrink noise features to zeros.

For each scenario, we randomly generate five sets of data. 
\begin{figure*}
        \centering
        \begin{subfigure}[b]{0.245\textwidth}
        \includegraphics[width=\textwidth]{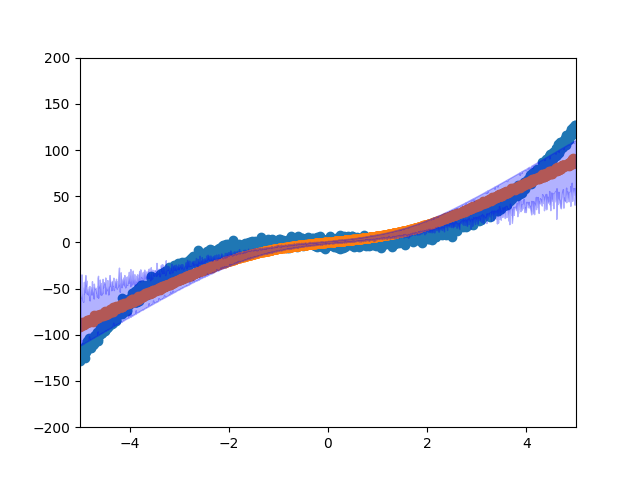}
        \caption{MC Dropout}
        \end{subfigure}
        \begin{subfigure}[b]{0.245\textwidth}
        \includegraphics[width=\textwidth]{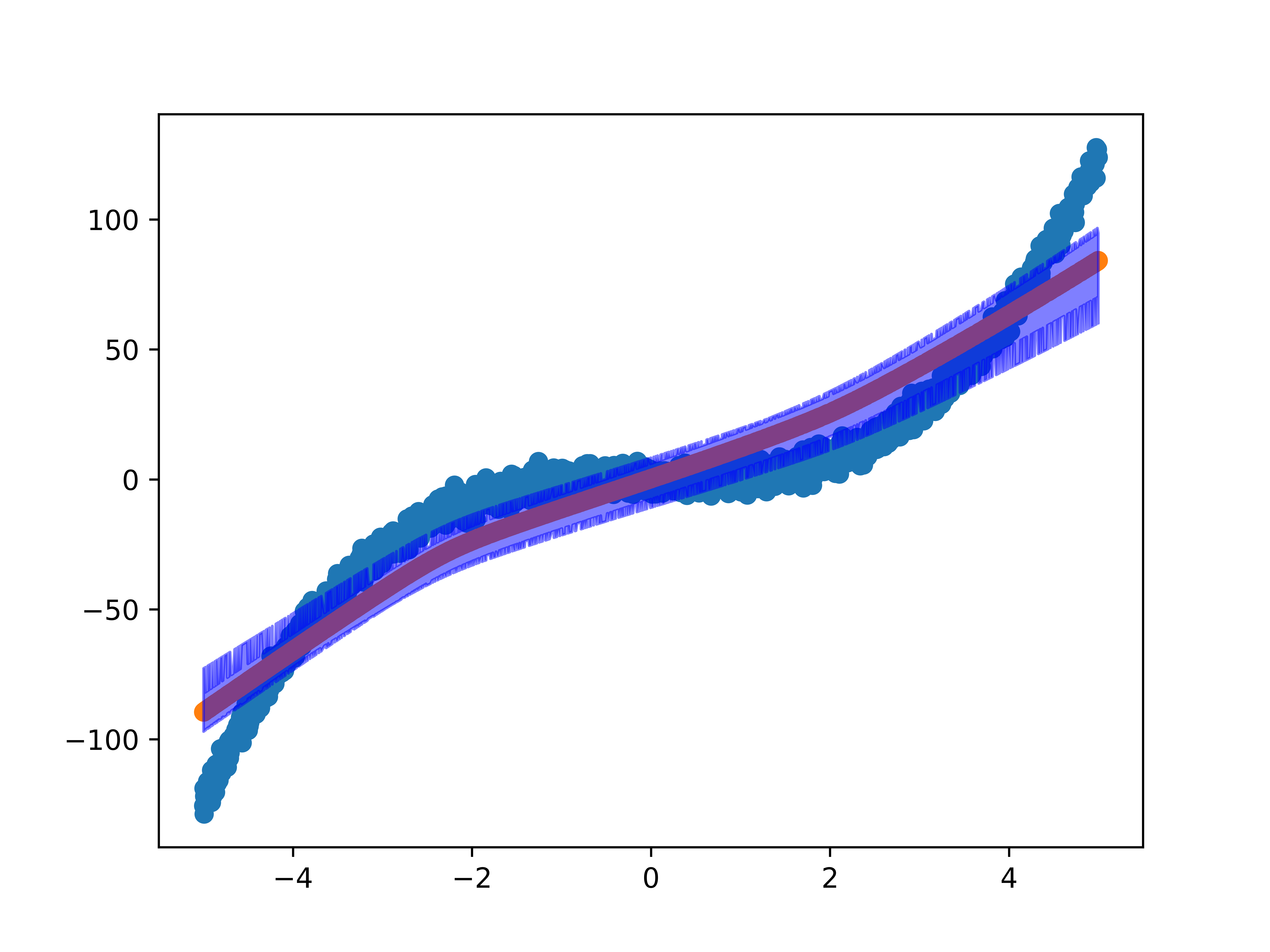}
        \caption{Gaussian BNN}
        \end{subfigure}
        \begin{subfigure}[b]{0.245\textwidth}
        \includegraphics[width=\textwidth]{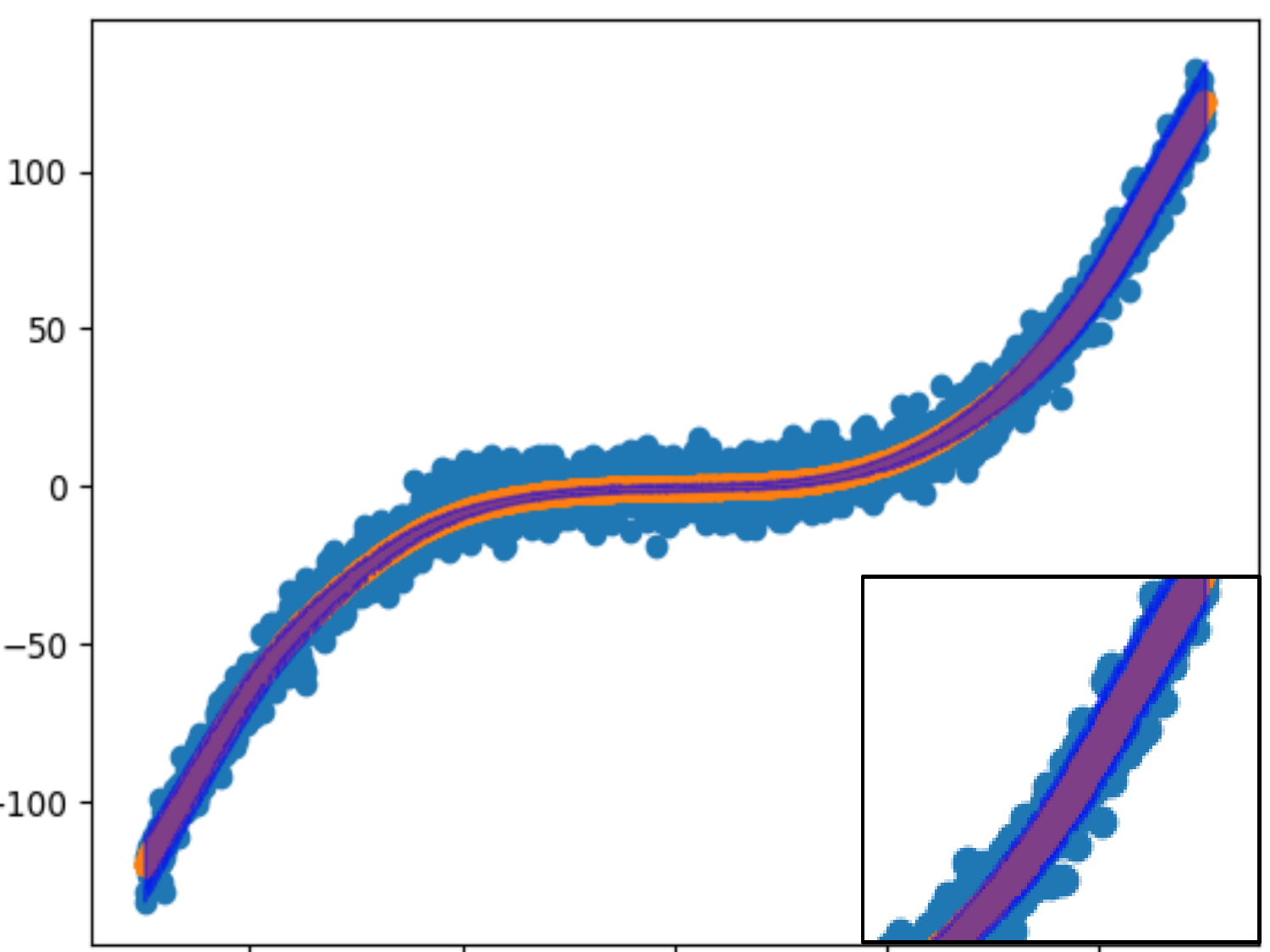}
        \caption{Horseshoe BNN}
        \end{subfigure}
        \begin{subfigure}[b]{0.245\textwidth}
        \includegraphics[width=\textwidth]{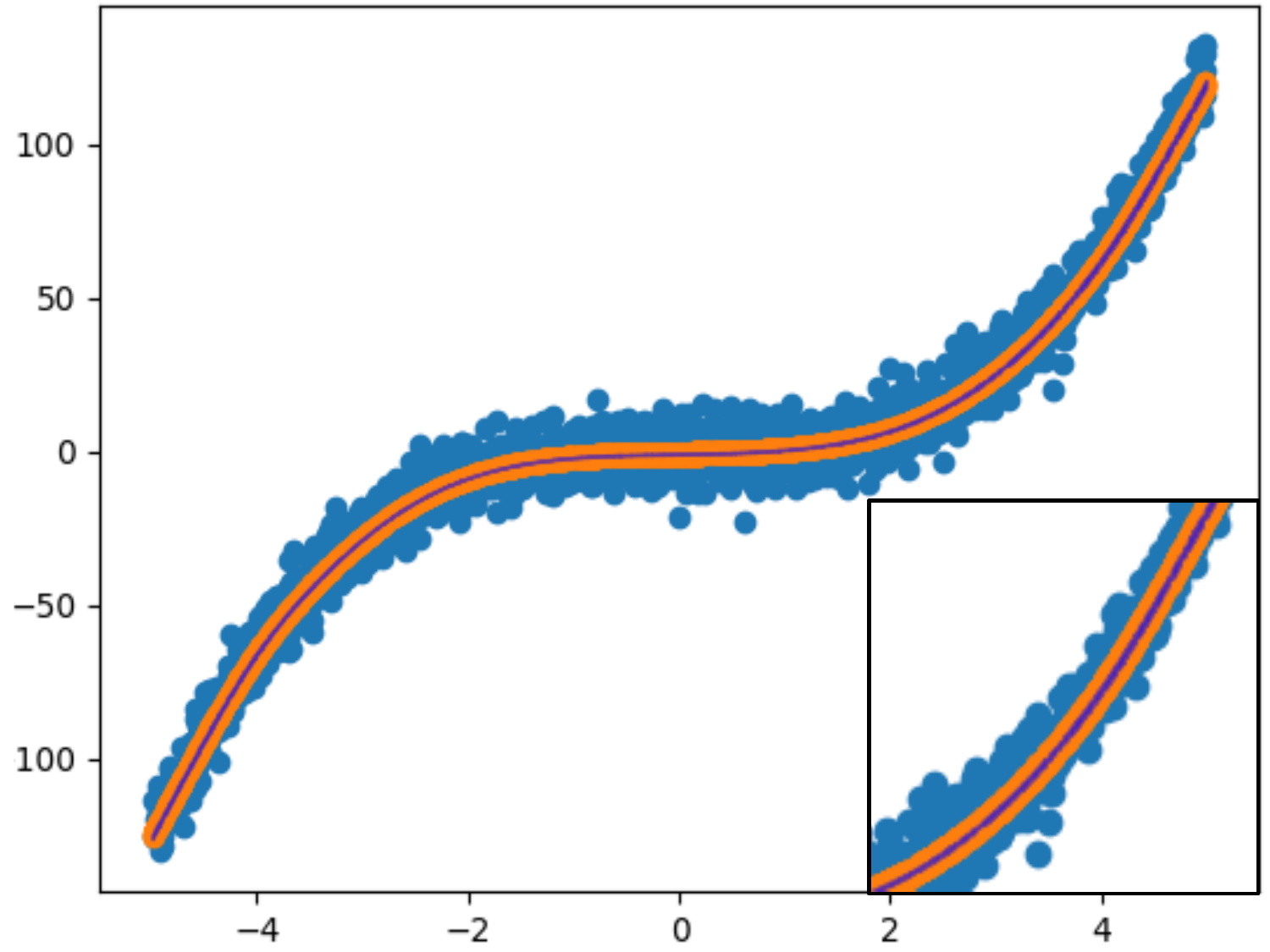}
        \caption{R2D2-Net}
        \end{subfigure}
        \caption{Prediction mean and confidence intervals of R2D2-Net at test time on $y_i = x_i^3 + \epsilon_i, \epsilon_i \sim \mathcal{N}(0, 9)$. The number of layers is 3 and the number of samples is 100 during the validation phase. 
        The blue dots are the ground truth data points, the yellow line is the mean of prediction and the blue shadow is the prediction interval. 
        We observe that the R2D2-Net yields a smaller prediction variance than MC Dropout, Gaussian BNN, and Horseshoe BNN.}
        \label{fig:r2d2_pred}
\end{figure*}
We split 80\% of the data as the training set and 20\% as the testing set. 
%
%
All methods are trained on 100 epochs and a batch size of 1024, with possible early stopping when the loss does not decrease for 5 epochs. 
For other priors included in the experiments, we report the performance with the best-performing optimization algorithms.

\para{Competitive Methods.}
We compare our method with a variety of existing BNN designs. 
The hyperparameter settings of each benchmark method and the summary of uncertainty measures used are presented in the supplementary materials.

We divide the baseline methods into combinations of priors and inference algorithms. 
We consider the following common priors for BNNs:
\begin{enumerate*}[label=(\arabic*)]
    \item  \textbf{Gaussian Prior (Gauss)}
    \item \textbf{Horseshoe Prior (HS)}
    \item \textbf{Spike-and-slab Prior (SaS)},
\end{enumerate*}
and the following inference algorithms:
\begin{enumerate*}[label=(\arabic*)]
    \item  \textbf{Stochastic Variational Inference (SVI) \cite{rudner2022FSVI}: } classical mean field variational inference which back-propagates the stochastic gradient on the BNN parameters;
    \item \textbf{Stochastic Gradient MCMC (SGMCMC) \cite{nemeth2021SGMCMC}};
    \item \textbf{Stochastic Gradient Langevin Dynamics (SGLD) \cite{welling2011SGLD}}.
\end{enumerate*}

\subsection{Experimental Results}

\para{Predictive Performance: R2D2-Net Achieves Competitive Performance with Deeper Layers.}
We compare the prediction MSE and variance of each BNN design. 
When $L = 0$, the model is equivalent to a linear regression. 
%
%
Table \ref{tab: results_simulation} presents the simulation results, and Figure \ref{fig:r2d2_pred} shows the prediction means and variances of R2D2-Net and the baseline BNN designs.

We observe that the R2D2-Net yields the smallest prediction error among all competitive designs, and overall a low inference error compared to the oracle truth.
%
\begin{table*}[t]

\caption{Simulation results on MSE and inference error under the R2D2-Net compared with different BNN designs and optimization algorithms on MLP with different numbers of layers $L=0, 1, 2, 3$. 
%
We report the mean of the ten replicates.
}
\centering
    \scalebox{1}{
        \begin{tabular}{llcccccccc p{1.2cm}}
        \toprule
        \multicolumn{10}{c}{\textbf{Non-Trivial Features}} \\ \hline
        \multicolumn{10}{c}{\textbf{Scenario 1: Polynomial Case}} \\ \hline
        \multicolumn{2}{l}{\textbf{}} &
        \multicolumn{2}{c}{\textbf{$L = 0$}} &
        \multicolumn{2}{c}{\textbf{$L = 1$}} &
        \multicolumn{2}{c}{\textbf{$L = 2$}} &
        \multicolumn{2}{c}{\textbf{$L = 3$}}
        \\
        \multicolumn{1}{l}{\textbf{Prior}} &
        \multicolumn{1}{l}{\textbf{Inf. Alg.}} & 
        \multicolumn{1}{c}{\textbf{MSE}} & 
            \multicolumn{1}{c}{\textbf{Inf. Err.}} &
        \multicolumn{1}{c}{\textbf{MSE}} & 
        \multicolumn{1}{c}{\textbf{Inf. Err.}} &
        \multicolumn{1}{c}{\textbf{MSE}} & 
        \multicolumn{1}{c}{\textbf{Inf. Err.}} &
        \multicolumn{1}{c}{\textbf{MSE}} & 
        \multicolumn{1}{c}{\textbf{Inf. Err.}} 
        \\ \hline
         \multirow{3}{*}{Gauss} & SVI & \textbf{363.22} & 1.19 & \textbf{9.4} & 369.86 & 9.46 & 1056.8 & 15.4 & 2720.93\\
         & SGMCMC & 363.27 & 1.17 & 9.61 & 452.32 & 18.64 & 1034.01 & 89.98 & 2555.65 \\
         & SGLD & 363.22 & 203.97 & 22.16 & 252.52 & 163.75 & 149.21 & 116.06 & 1021.6 \\ \hline 
         \multirow{3}{*}{SaS} & SVI & 1432.12 & 0.43 & 213.79 & 122.4 & 36.1 & 535.9 & 49.89 & 905.75\\
         & SGMCMC & 1364.4 & 0.13 & 131.28 & 153.58 & 94.84 & 434.9 & 145.85 & 718.72 \\
         & SGLD & 370.54 & 174.1 & 320.98 & 65.28 & 168.11 & 110.75 & 44.65 & 336.49\\ \hline
        \multirow{2}{*}{HS} &SVI & 862.26 & 2.47 & 313.72 & \textbf{19.44} & 113.35 & 445.99 & 22.98 & 552.67 \\
         & SGMCMC &  361.7 & 2.68 & 50.48 & 32.73 & 119.57 & 217.48 & 142.48 & 574.41\\ \hline
        \multirow{3}{*}{\textbf{R2D2}} & SGMCMC & 378.91 & 142.8 & 18.18 & 101.83 &  37.05 & 164.00 & 46.62 & 361.11\\
         & SGLD  & 358.41 & 2.79 & 14.66 & 77.72 & 29.3 & 170.89 & 20.61 & 342.5\\
         & \textbf{SVGI} &  414.36& \textbf{0.10} & 10.23 & 50.28 & \textbf{9.18} & \textbf{103.48} & \textbf{8.81} &\textbf{ 332.67} \\ \hline
        \multicolumn{10}{c}{\textbf{Scenario 2: Low-dimensional Non-linear Regression}} \\ \hline
        \multicolumn{2}{l}{\textbf{}} &
        \multicolumn{2}{c}{\textbf{$L = 0$}} &
        \multicolumn{2}{c}{\textbf{$L = 1$}} &
        \multicolumn{2}{c}{\textbf{$L = 2$}} &
        \multicolumn{2}{c}{\textbf{$L = 3$}}
        \\
        \multicolumn{1}{l}{\textbf{Prior}} &
        \multicolumn{1}{l}{\textbf{Inf. Alg.}} & 
        \multicolumn{1}{c}{\textbf{MSE}} & 
        \multicolumn{1}{c}{\textbf{Inf. Err.}} &
        \multicolumn{1}{c}{\textbf{MSE}} & 
        \multicolumn{1}{c}{\textbf{Inf. Err.}} &
        \multicolumn{1}{c}{\textbf{MSE}} & 
        \multicolumn{1}{c}{\textbf{Inf. Err.}} &
        \multicolumn{1}{c}{\textbf{MSE}} & 
        \multicolumn{1}{c}{\textbf{Inf. Err.}} 
        \\ \hline
        \multirow{3}{*}{Gauss} & SVI & 540.2 & 0.42 & 24.03 & 417.64 & 10.74 & 1279.85 & 12.57 & 2726.73 \\
         & SGMCMC & 540.17 & 0.42 & 274.17 & 242.59 & 19.6 & 1109.39 & 181.44 & 2765.65\\
        & SGLD & 552.66 & 75.25 & 400.31 & 92.34 & 55.81 & 235.75 & 64.71 & 494.19 \\ \hline
            \multirow{3}{*}{SaS} & SVI & 814.26 & 0.9 & 459.42 & 73.19 & 28.77 & 438.81 & 30.12 & 751.85 \\
         & SGMCMC & 754.62 & 1.86 & 468.78 & 66.36 & 44.7 & 392.22 & 51.98 & 428.58 \\
        & SGLD & 547.05 & 62.87 & 499.41 & 44.92 & 420.5 & 101.58 & 90.91 & 350.62\\ \hline
        \multirow{2}{*}{HS} & SVI & 560.73 & 4.85 & 435.69 & 34.06 & 136.51 & 265.6 & 31.3 & 479.22\\
         & SGMCMC & 546.9 & 3.13 & 322.29 & \textbf{32.46} & 102.57 & 172.64 & 118.96 & 363.33\\ \hline
        \multirow{3}{*}{\textbf{R2D2}} & SGMCMC & 509.29 & 4.48 & 35.09 & 34.66 & 93.42 & 249.92 &  130.47 & 513.81 \\
         & SGLD & 509.38 & 0.31 & 31.50 & 63.14 & 55.44 & 183.46 & 52.56 & 352.72\\
         & \textbf{SVGI} & \textbf{509.23} & \textbf{0.26} & \textbf{22.38} & 46.48 & \textbf{10.26} & \textbf{84.52} & \textbf{12.24} & \textbf{348.51} \\\hline
        \multicolumn{10}{c}{\textbf{Trivial Features}} \\ \hline
        \multicolumn{10}{c}{\textbf{Scenario 3: High-dimensional Non-linear Regression}} \\ \hline
        \multicolumn{2}{l}{\textbf{}} &
        \multicolumn{2}{c}{\textbf{$L = 0$}} &
        \multicolumn{2}{c}{\textbf{$L = 1$}} &
        \multicolumn{2}{c}{\textbf{$L = 2$}} &
        \multicolumn{2}{c}{\textbf{$L = 3$}}
        \\
        \multicolumn{1}{l}{\textbf{Prior}} &
        \multicolumn{1}{l}{\textbf{Inf. Alg.}} & 
        \multicolumn{1}{c}{\textbf{MSE}} & 
        \multicolumn{1}{c}{\textbf{Inf. Err.}} &
        \multicolumn{1}{c}{\textbf{MSE}} & 
        \multicolumn{1}{c}{\textbf{Inf. Err.}} &
        \multicolumn{1}{c}{\textbf{MSE}} & 
        \multicolumn{1}{c}{\textbf{Inf. Err.}} &
        \multicolumn{1}{c}{\textbf{MSE}} & 
        \multicolumn{1}{c}{\textbf{Inf. Err.}} 
        \\ \hline
        \multirow{3}{*}{Gauss} & SVI & 5.52 & 32.5 & 5.82 & 6948.52 & 5.57 & 8648.4 & 5.58 & 10519.14 \\
         & SGMCMC & 6.04 & 33.32 & 7.21 & 9715.69 & 5.84 & 9605.35 & 7.12 & 10368.32\\
        & SGLD & 5.39 & 12.32 & 6.13 & 628.15 & 4.16 & 789.62 & 4.07 & 979.67\\ \hline
            \multirow{3}{*}{SaS} & SVI & 5.23 & 19.58 & 4.81 & 1100.96 & 4.62 & 1543.72 & 4.41 & 1803.17\\
         & SGMCMC & 5.24 & 20.24 & 5.3 & 1328.69 & 5.44 & 1138.89 & 5.15 & 1320.50\\
         & SGLD & 7.33 & 12.45 & 4.65 & 841.21 & 4.17 & 905.34 & 4.14 & 1120.53\\ \hline
        \multirow{2}{*}{HS} & SVI &  4.68 & 14.32 & 8.79 & 649.24 & 7.94 & 814.56 & 5.99 & 1023.07 \\
         & SGMCMC & 4.49 & 36.05 & 6.04 & 1447.7 & 6.43 & 623.02 & 6.25 & 2009.38 \\ \hline
        \multirow{3}{*}{\textbf{R2D2}} & SGMCMC  & 7.1 & 11.71 & 8.51 & 711.03 & 4.61 & 602.83 & 4.89 & 942.12\\
         & SGLD & 4.68 & 11.63 & 4.83 & 532.27 & 6.00 & 547.56 & 4.16 & 787.50\\
         & \textbf{SVGI}& \textbf{4.35} & \textbf{11.47} & \textbf{4.63} & \textbf{480.26} & \textbf{4.15} & \textbf{510.87} & \textbf{4.06} & \textbf{719.15}\\
        \bottomrule
      \end{tabular}}
    \label{tab: results_simulation}
{\scriptsize \begin{spacing}{0.5}
SaS: spike-and-slab prior; SVI: stochastic variational inference; SGMCMC: stochastic gradient MCMC; SGLD: stochastic gradient langevin dynamics; SVGI: stochastic variational Gibbs inference.
\end{spacing}
}

\vspace{-3mm}
\end{table*}

The R2D2-Net also shows greater improvement in prediction performance (i.e., smaller prediction MSE) as the number of layers increases.
This demonstrates that the R2D2-Net is more capable of supporting deeper BNN architectures than other BNN designs.
On the other hand, when compared to traditional sampling-based optimization algorithms, we observe that using the Gibbs posterior as the variational posterior would lead to more stable convergence performance and better predictive results. 
This empirically demonstrates the effectiveness of using the Gibbs posterior in variational inference.
This highlights the variance inflation issues of vanilla BNNs.

\para{Shrinkage Performance: R2D2 Prior Best Shrinks Unnecessary Neurons to Zero.}
We study the shrinkage performance of R2D2-Net in comparison with the Horseshoe BNN \cite{ghosh2019horseshoebnn}.
We investigate the distribution of the smallest node weight vectors to compare the shrinkage performance among different priors. 
We plot the distributions of coefficients $w_{jl}$ with the smallest magnitude $\| \mathbb{E}[w_{jl}]\|$ (Figure \ref{fig: denties}).
We observe that the weight samples of the R2D2-Net have the highest concentration rate at zero compared with Horseshoe BNN and Gaussian BNN. 
This validates that the highest concentration rate property of the R2D2 prior also holds when generalized to neural networks. 
We also validate that the R2D2-Net has the best shrinkage performance than existing BNNs with other priors.

\noindent
\para{Inference Accuracy.}
We also evaluate the inference accuracy of R2D2-Net in addition to predictive accuracy.
We consider a two-layer MLP (multi-layer perceptron) and use the HMC algorithm to generate the oracle truth posteriors of the BNN parameters. 
We use the L2-normalized difference between the learned parameters and the oracle truth as the inference error. 
%

\begin{figure*}
        \centering
        \includegraphics[width=0.9\textwidth, height=0.25\textwidth]{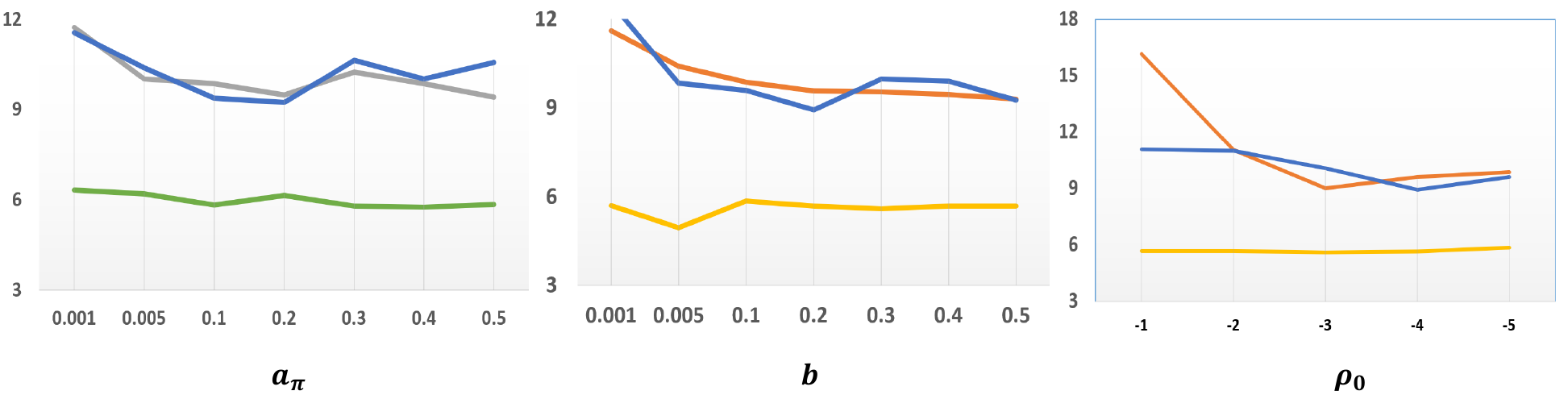}
        \caption{Ablation studies of our method to different hyperparameters.
        We run the three simulation scenarios (S1--S3) with an R2D2 MLP with $L = 3$, and report the testing MSEs with respect to different values of hyperparameters $a_\pi$ (left), $b$ (middle), $\rho_0$ (right).
        }
        \label{fig: ablation_hyperparameters}
\end{figure*}

\begin{figure}
        \centering
        \includegraphics[width=0.5\textwidth]{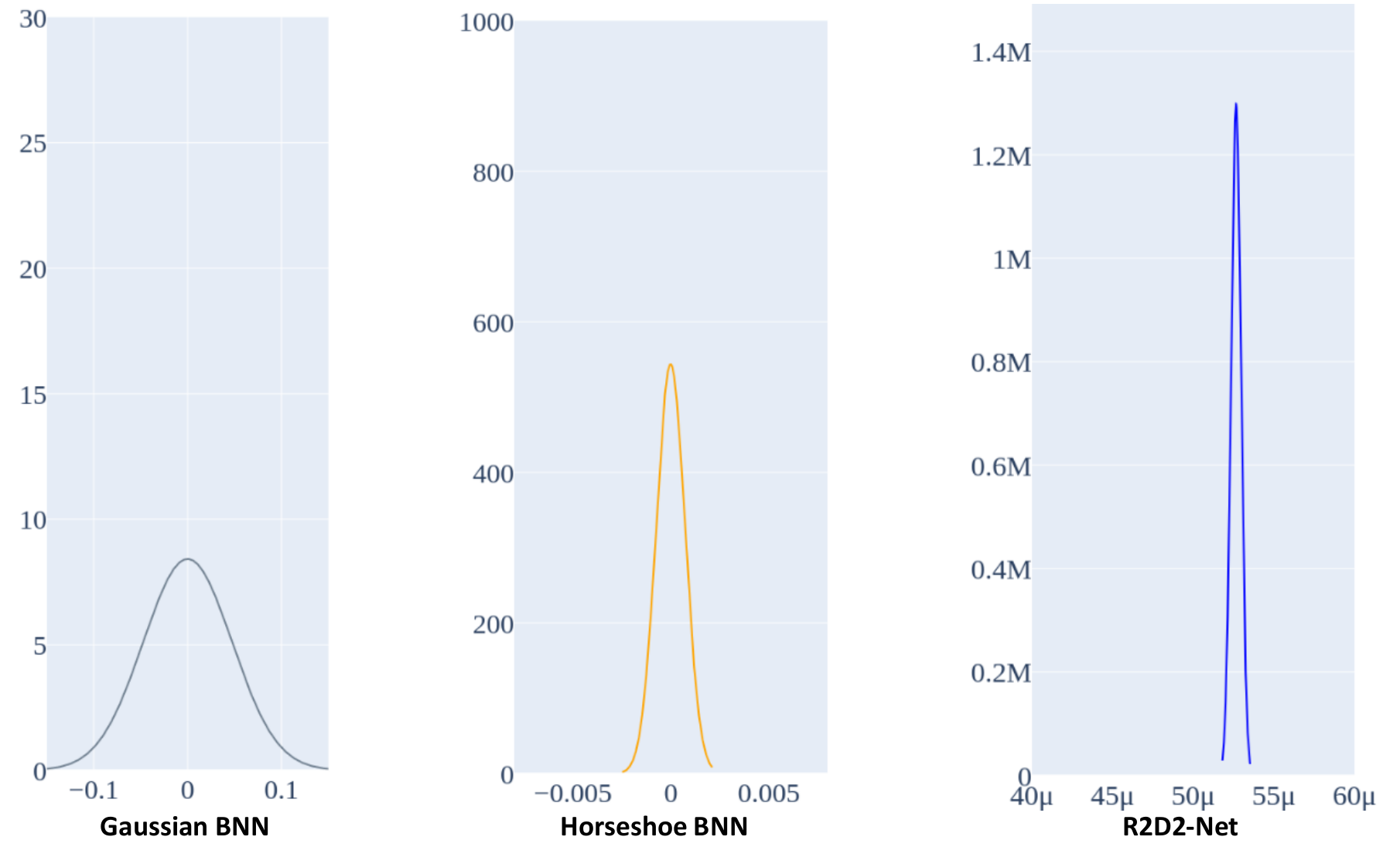}
        \caption{Density plots of the weight samples of Gaussian BNN,  Horseshoe BNN, and R2D2-Net.
        We choose the weights that have the least magnitude from the first layer of a three-layer MLP.
        We observe that R2D2-Net has the highest concentration rate at zero.
        }
        \label{fig: denties}
\end{figure}

\para{Impact of Hyperparameters. }
We investigate how sensitive the R2D2-Net is to the changes in the hyperparameters, such as $a_\pi, b$ and $\rho_0$. 
We perform the evaluation using the simulation scenarios in Section \ref{sec: simulations}.
Figure \ref{fig: ablation_hyperparameters} presents the results using an R2D2 MLP with $L=3$.
We observe that our method is robust to changes in these hyperparameters. 
The performance of R2D2-Net is more sensitive to the prior variance parameter $\rho_0$ than the other hyperparameters $a_\pi$ and $b$ of the R2D2 prior in Eq. (\ref{eq: r2d2_prior}).  

\section{Experiments on Real Data}
%
%

We further validate the capability of R2D2-Net with real datasets (i.e., TinyImageNet) and larger architectures (i.e.,  residual nets).
\vspace{-3mm}
\subsection{Experimental Setup}
\para{Datasets.}
We evaluate the R2D2-Net on standard computer vision datasets in comparison with existing methods. 
Table \ref{tab:Datasets summary} provides a summary of the datasets.
For \textbf{image classification}, we use CIFAR 10, CIFAR 100, and TinyImageNet as the benchmark datasets.
We perform 5-fold cross-validation to evaluate each method. 
We use accuracy, macro F1 score, and area under the receiver operating curve (AUROC) as the evaluation metric, and report the mean and standard deviation of each metric. 
For \textbf{uncertainty estimation}, we assess the performance of the neural networks using the out-of-distribution (OOD) detection task, with AUROC and the area under the precision-recall curve (AUPR) as the evaluation metrics. 
We treat the images in the CIFAR 10 \cite{krizhevsky2009CIFAR} dataset as the in-distribution data and the images from the fashion MNIST, OMNIGLOT, and SVHN \cite{xiao2017/fashionminst} as the OOD samples. 
In contrast to some existing approaches \cite{malinin2018DPN, sensoy2018EDL}, we train the classifier with in-distribution data only (i.e., the classifier does not see the OOD data during training).

\begin{table}[!ht]
\caption{Summary of Datasets}
    \centering
    \scalebox{0.9}{
\begin{tabular}{lcccc}
\toprule
\multicolumn{1}{l}{\textbf{Datasets}} &
\multicolumn{1}{c}{\textbf{No. Classes}} &
\multicolumn{1}{c}{\textbf{No. Training}} &
\multicolumn{1}{c}{\textbf{No. Testing}}\\ \hline
\multicolumn{1}{l}{MNIST} & 10  & 60,000 & 10,000 \\ \hline
\multicolumn{1}{l}{Fashion-MNIST} & 10  & 60,000 & 10,000\\ \hline
\multicolumn{1}{l}{OMNIGLOT} & 50  & 13,180 & 19,280\\ \hline
\multicolumn{1}{l}{SVHN} & 10  & 73,257 & 26,032\\ \hline
\multicolumn{1}{l}{CIFAR-10} &  10 & 60,000 & 10,000\\ \hline
\multicolumn{1}{l}{CIFAR-100} &  100 & 60,000 & 10,000\\ \hline
\multicolumn{1}{l}{TinyImageNet} &  200 & 80,000 & 20,000\\ \hline
\multicolumn{1}{l}{DRD} &  2 & 50& 100\\ 
\bottomrule
\end{tabular}}
\label{tab:Datasets summary}
\end{table}

\para{Competitive Methods.}
Additionally, we consider a series of classical neural network designs for comparison:
\begin{enumerate*}[label=(\arabic*)]
\item \textbf{Frequentist CNN} (Freq): the original frequentist neural network architecture;
\item \textbf{MC Dropout} \cite{gal2016mcdropout} (MCD): using repeated dropouts on trained weights to draw Monte Carlo samples of the weights of the BNNs (reproduced from \cite{gal2016mcdropout});
\item \textbf{RadialBNN} \cite{farquhar2020radialBNN} (Radial): sampling from the hyperspherical coordinate system to resolve the problem in the original MFVI where the probability mass is far from the true mean;
and it is implemented as a comparable optimization algorithm with the Gaussian prior; 
\item \textbf{Deep Ensembles} \cite{lakshminarayanan2017deepensembles} (DE): it uses a finite ensemble of deep neural networks to approximate the posterior weight distribution.
\end{enumerate*}

In addition to existing BNN designs, we add two entropy-based uncertainty estimation methods for comparison. 
Because we adopt entropy as the uncertainty metric (as this is a classic metric for classification uncertainty), the OOD performance may be slightly worse than their respective state-of-the-art (SOTA) performances: 
(1) \textbf{DPN} \cite{malinin2018DPN}: it assumes a Dirichlet distribution on the classification output and trains an OOD classifier by minimizing the KL divergence between the prior and posterior distributions;
(2) \textbf{EDL} \cite{sensoy2018EDL}: in addition to DPN \cite{malinin2018DPN}, EDL trains the classifier with the cross-entropy loss and the KL divergence between the prior and posterior distributions.

\subsection{Image Classification: R2D2 Shrinkage Improves Predictive Performance}

Table \ref{tab: CIFAR_Alex} presents the image classification results of our R2D2-Net in comparison with existing methods. 
\begin{table*}[h]

\caption{Image classification results [\%] of our proposed method on CIFAR 10 and CIFAR 100 with the AlexNet \cite{krizhevsky2012alexnet}. 
Standard deviations are shown in brackets. 
\textbf{Boldface} represents the best performance among BNN designs, while * represents the best performance among all models.
}
\setlength{\tabcolsep}{2mm}
    \centering
    \scalebox{0.9}{
        \begin{tabular}{ll|cc|cc|cc p{1.3cm}}
        \toprule
        \multicolumn{2}{l}{\textbf{}} &
        \multicolumn{2}{c}{\textbf{CIFAR 10}} &
        \multicolumn{2}{c}{\textbf{CIFAR 100}} &
        \multicolumn{2}{c}{\textbf{TinyImageNet}}
        \\
        \multicolumn{1}{l}{\textbf{Prior}} &
        \multicolumn{1}{l}{\textbf{Inference Alg.}} &
        \multicolumn{1}{c}{\textbf{AUROC}} & 
        \multicolumn{1}{c}{\textbf{Accuracy}} &
        \multicolumn{1}{c}{\textbf{AUROC}} &
        \multicolumn{1}{c}{\textbf{Accuracy}} & 
        \multicolumn{1}{c}{\textbf{AUROC}} & 
        \multicolumn{1}{c}{\textbf{Accuracy}} 
        \\ \hline
        Freq &  & 92.70 (1.5)* & 65.03 (1.4)  & 90.95 (0.2) & 31.05 (0.4)& 88.37 (1.3) & 18.30 (0.4)\\ \hline
        MCD & SVI & 90.09 (0.2) & 55.08 (0.6) & 87.67 (1.3)& 21.92 (1.1) & 86.23 (1.7) & 17.28 (1.5) \\ 
        DE &  & 90.67 (0.6) & 62.41 (0.5) & 87.97 (0.9) & 24.63 (0.4) & 86.25 (0.4) & 13.73 (0.5) \\ \hline
         \multirow{5}{*}{Gauss} & SVI & 91.37 (1.2) & 60.28 (1.5) & 87.24 (1.2) & 23.6 (0.6) & 87.64 (0.2) & 16.82 (0.9)\\
         & MFVI&  91.11 (0.9) & 59.27 (1.1) & 87.69 (0.9) & 23.06 (0.2) & 86.01 (0.3) & 12.78 (0.6)\\
        & Radial & 91.22 (0.8) & 63.24 (0.8) & 89.20 (1.0) & 25.70 (0.5) & 84.35 (0.6) & 12.12 (0.5)\\
        & SGMCMC & 83.79  (0.9) & 41.91 (0.8) & 78.71 (1.0) & 10.32 (1.2) &  84.37 (0.3) & 10.19 (0.5) \\
         & SGLD & 89.34 (1.3) & 52.22 (1.0) & 79.21 (1.2) & 12.31 (1.3) & 84.42 (0.3) & 10.15 (0.6)\\ \hline
        \multirow{3}{*}{SaS} & SVI & 91.69 (0.4) & 60.84 (0.7) & 89.68 (0.1) & 29.01 (0.2) & 85.37 (0.2) & 14.49 (0.3) \\
         & SGMCMC &  84.10 (0.3) & 43.03 (0.4) & 76.44 (0.8) & 8.35 (1.2) & 76.74 (1.1) & 4.54 (1.5)\\
         & SGLD & 86.14 (0.2) & 45.13 (0.3) & 80.74 (0.1) & 9.89 (0.2) & 71.27 (2.3) & 3.87 (1.4)\\ \hline
        \multirow{2}{*}{HS} & SVI & 91.99 (0.8) & 65.01 (0.3)  & 91.37 (0.2) &  33.27 (0.3) & 88.71 (2.0) & 20.33 (1.2) \\
        & SGMCMC & 89.71 (0.5) & 54.96 (0.8) & 80.12 (0.1) & 14.92 (0.3) & 85.71 (1.3) & 14.08 (1.1) \\ \hline
        \multirow{2}{*}{R2D2} & SGMCMC &   88.36 (0.3) & 55.37 (0.4) & 85.48 (0.1) & 20.18 (0.3) & 77.83 (0.9)  & 6.17 (1.2)\\
         & \textbf{SVGI} & \textbf{92.49 (0.2)} & \textbf{65.10 (0.02)}*  & \textbf{92.48 (0.03)}* & \textbf{36.12 (0.5)}* &
        \textbf{88.76 (0.5)*} &
        \textbf{20.55 (0.4)*}\\
        \bottomrule
        \end{tabular}}
    \label{tab: CIFAR_Alex}

\end{table*}
We assess our method on standard image classification benchmarks --- CIFAR 10, CIFAR 100, and TinyImageNet.
We fix the model architecture as AlexNet \cite{krizhevsky2012alexnet} for fair comparison.
\begin{figure}
    \centering
    \includegraphics[width=0.4\textwidth]{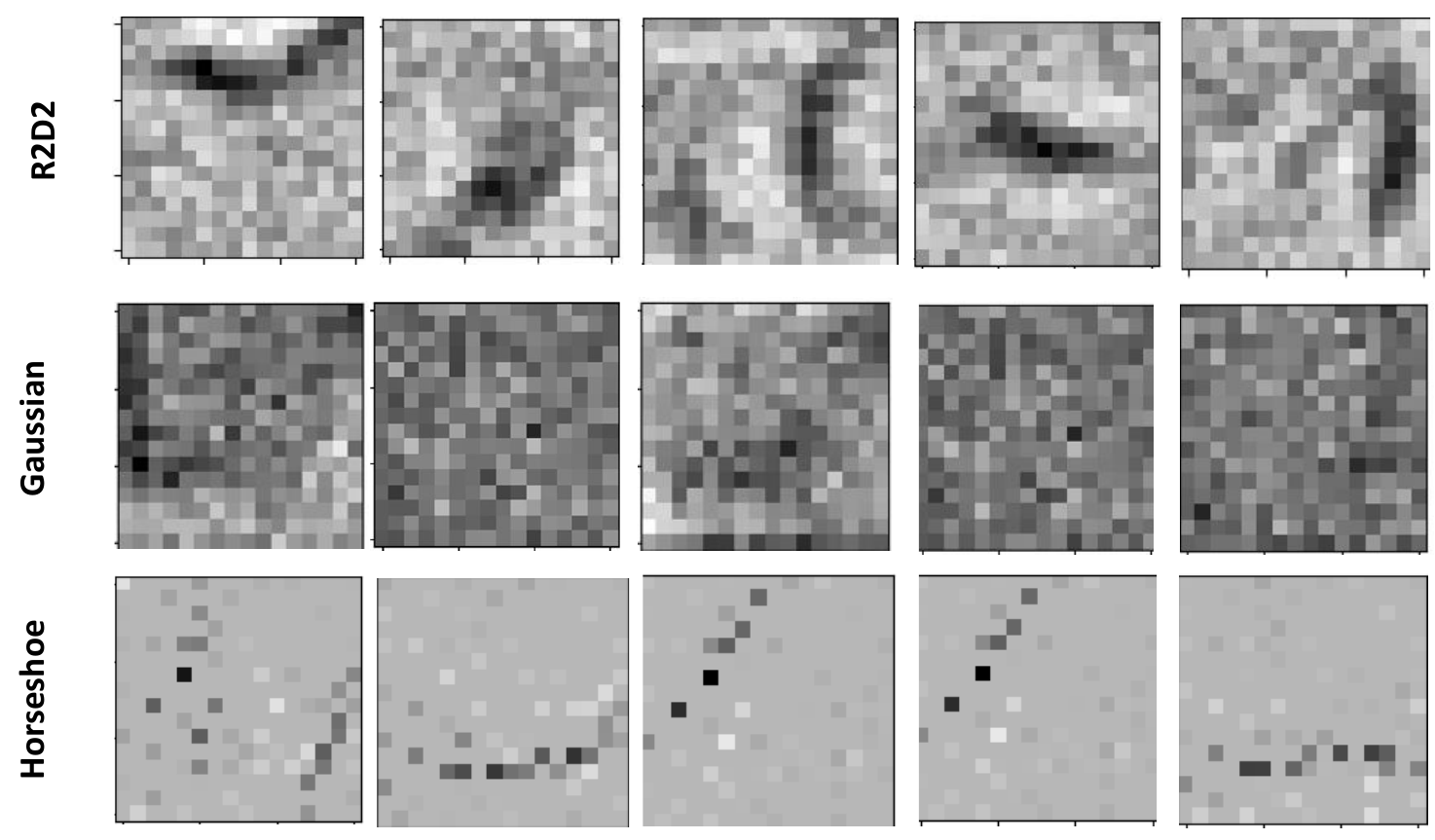}
    \caption{Five largest-norm convolutional filters of the R2D2-Net, Gaussian BNN, and Horseshoe BNN.  
    We use a simple CNN with one convolutional layer and one linear layer for illustrative purposes.
    }
    \label{fig: filters}
\end{figure}
%
%
Not only does our proposed method outperform the existing BNN designs, but it also occasionally outperforms the frequentist design. 
It is noteworthy that since BNNs impose a natural regularization on the weights, it is difficult for BNN designs to outperform their frequentist counterpart. 
This demonstrates that choosing the R2D2 prior can potentially lead to the best variable selection outcome. 
The R2D2 prior can select a suitable subset of weights with its shrinkage properties, while the frequentist design cannot.
Hence, its prediction performance can be more satisfactory than the original frequentist design. 
On the other hand, from the standard deviations,
we observe that adopting the Gibbs posterior of the R2D2 prior in variational inference would lead to more consistent and stable convergence results than other optimization algorithms.

%
%
We visualize the five largest-norm filters of the Gaussian BNN, Horseshoe BNN, and the R2D2-Net to compare their capabilities to select features (Figure \ref{fig: filters}).
We observe that the largest filters of Gaussian BNN  can capture the pattern of the image, while the noise is heavy. 
On the contrary, 
the Horseshoe BNN shrinks noise more effectively, while it  suffers from feature losses (i.e., over-shrinkage).
Compared to both above, the filters of R2D2-Net have less noise, while preserving most of the features.
%
Since the R2D2 prior has heavier tails than the Horseshoe prior, it can preserve large signals in the filter weights and avoid over-shrinkage, as demonstrated by the difference in filter patterns in Figure \ref{fig: filters}.

\subsection{Uncertainty Estimation: R2D2 Shrinkage Captures Important Variance.}

\begin{figure}
    \centering
    \includegraphics[width=0.4\textwidth]{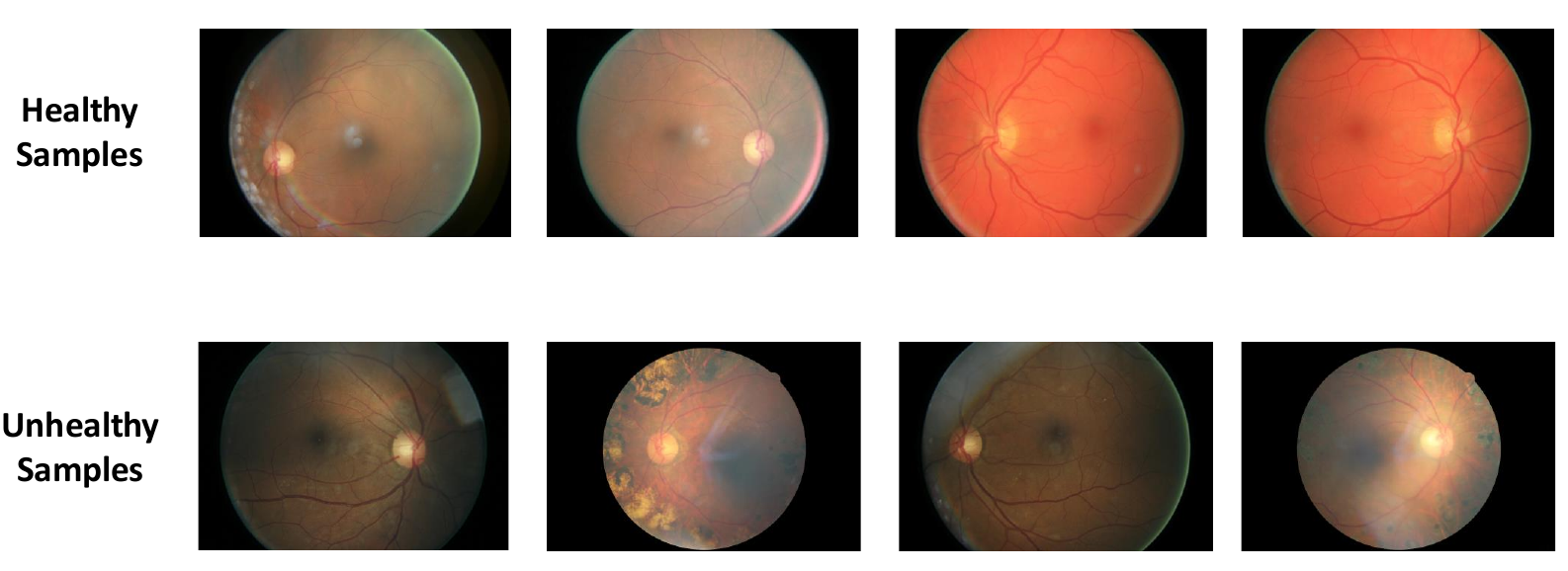}
    \caption{In-distribution (healthy) and OOD (unhealthy) samples detected by our method. 
    }
    \label{fig: DRD detection}
    \vspace{-4mm}
\end{figure}

We further compare the performance of uncertainty estimation with the existing BNN designs. 
We additionally include two entropy-based uncertainty estimation methods: DPN \cite{malinin2018DPN} and EDL \cite{sensoy2018EDL}, which estimate uncertainties based on the assumption of Dirichlet distribution on latent probabilities. 
We use the classification entropy as the uncertainty measure. 
%
The entropy of classification is defined as
    \begin{align*}
        \mathbb{H}[p(\bm \mu|\mathcal{D})] = - \sum_{c=1}^K p(\mu_c| \mathcal{D}) \log p(\mu_c| \mathcal{D}), 
\end{align*}
    where $P(\mu_c|\mathcal{D})$ is the predictive probability of class $c$, and $K$ is the number of classes for classification.
More information on baseline methods and uncertainty measures is given in the supplementary materials.
We adopt the OOD detection task to evaluate the performance of the R2D2-Net for estimating the uncertainty in the data. 
The OOD detection aims to identify whether the input data are in-distribution or from a different dataset.
Tables \ref{tab: results_cifarOOD} and \ref{tab: results_mnistOOD}  present the AUROC and the AUPR of the R2D2-Net using the classification entropy as the uncertainty measure.
We treat MNIST and CIFAR 10 as the in-distribution datasets and FasionMNIST, OMNIGLOT, and SVHN as the OOD datasets.
We also include a medical image benchmark, Diabetic Retinopathy Detection (DRD) dataset, where we treat healthy samples as in-distribution data and unhealthy samples as the OOD dataset.
Examples of the healthy and unhealthy samples detected can be found in Figure \ref{fig: DRD detection}.
%
It is clear that our R2D2-Net demonstrates a satisfactory performance over the baseline methods.
This shows that using an R2D2 prior on the weights can effectively shrink the noises in parameters while maintaining a non-trivial variance structure. 
These preserved variances can be used to represent the model-wise uncertainties. 
Hence, the R2D2-Net can produce more accurate uncertainty estimates than existing Bayesian and non-Bayesian approaches.

\begin{table*}[t]
\caption{The OOD detection performance of the R2D2-Net compared with various BNN designs under the LeNet \cite{lecun1989LeNet}, using CIFAR 10 as the in-distribution dataset.
%
%
The best performance among all methods is highlighted in boldface.
}
    \centering
    \scalebox{0.9}{
        \begin{tabular}{lcc|cc|cc p{1.2cm}}
        \toprule
        \multicolumn{1}{l}{\textbf{}} &
        \multicolumn{2}{c}{\textbf{Fashion MNIST}} &
        \multicolumn{2}{c}{\textbf{OMNIGLOT}} &
        \multicolumn{2}{c}{\textbf{SVHN}}
        \\
        \multicolumn{1}{l}{\textbf{Models/Priors}} &
        \multicolumn{1}{c}{\textbf{AUROC}} & 
        \multicolumn{1}{c}{\textbf{AUPR}} &
        \multicolumn{1}{c}{\textbf{AUROC}} & 
        \multicolumn{1}{c}{\textbf{AUPR}} &
        \multicolumn{1}{c}{\textbf{AUROC}} & 
        \multicolumn{1}{c}{\textbf{AUPR}} 
        \\ \hline
        DE & 71.02 & 76.81 & 86.77 & 90.35 & 61.01 & 62.59\\
        MCD & 81.80 & 74.22 & 80.03 & 82.06 & 68.58 & 78.53 \\ 
        Gauss-SVI & 74.49 & 86.78 & 78.58 & 81.82 & 70.57 & 79.30 \\
        Gauss-MFVI & 85.45 & 79.55 & 89.17 & 91.64 & 76.02 & 85.08 \\
        Gauss-Radial & 83.86 & 81.26 & 75.39 & 74.91 & 67.74 & 81.91\\
        HS-SVI & 80.76 & 75.99 & 86.66 & 90.18 & 70.11 & 78.39\\
        SaS-SVI & 91.55 & 88.89 & 89.03 & 88.64 & 78.25 & 85.11 \\
        DPN  & 87.07 & 83.75 & 87.07 & 83.75 & 57.48 & 77.76\\ 
        EDL & 89.26 & 86.16 & 66.53 & 67.12 & 69.57 & 83.74\\
        \textbf{R2D2-Net}  & \textbf{92.85}  &  \textbf{94.09} & \textbf{91.95} & \textbf{92.25}& \textbf{79.84} & \textbf{89.24}\\
        \bottomrule
        \end{tabular}
    }
    \label{tab: results_cifarOOD}
\end{table*}

\begin{table*}[t]
\caption{The OOD detection performance of the R2D2-Net compared with various BNN designs under the LeNet \cite{lecun1989LeNet}, using MNIST as the in-distribution dataset.
%
%
The best performance among all methods is highlighted in boldface.}
    \centering
    \scalebox{0.9}{
        \begin{tabular}{lcc|cc|cc p{1.2cm}}
        \toprule
        \multicolumn{1}{l}{\textbf{}} &
        \multicolumn{2}{c}{\textbf{Fashion MNIST}} &
        \multicolumn{2}{c}{\textbf{OMNIGLOT}} &
        \multicolumn{2}{c}{\textbf{SVHN}}
        \\
        \multicolumn{1}{l}{\textbf{Models}} &
        \multicolumn{1}{c}{\textbf{AUROC}} & 
        \multicolumn{1}{c}{\textbf{AUPR}} &
        \multicolumn{1}{c}{\textbf{AUROC}} & 
        \multicolumn{1}{c}{\textbf{AUPR}} &
        \multicolumn{1}{c}{\textbf{AUROC}} & 
        \multicolumn{1}{c}{\textbf{AUPR}} 
        \\ \hline
        DE & 90.70 & 91.08 & 99.70 & 91.08 & 99.21 & 99.68\\
        MCD & 81.80 & 74.22 & 80.03 & 82.06 & \textbf{99.96} & \textbf{99.96} \\ 
        Gauss-SVI & 98.36 & 98.36 & 99.17 & 99.38  & 98.95 & 99.10 \\
        Gauss-MFVI & 98.52 & 98.48& 98.94 & 99.11 & 99.91 & \textbf{99.96} \\
        Gauss-Radial & 98.2 & 97.94 & 98.52 & 98.73 & 99.64 & 99.85\\
        HS-SVI & 80.76 & 75.99 & 99.06 & 99.65 & 99.35 & 98.74\\
        SaS-SVI & 98.28 & 98.16 & 99.57 & 99.57 & 99.90 & 99.96\\
        DPN  & 98.70 & 98.80 & 99.96 & 99.96 & \textbf{99.96} & \textbf{99.96}\\ 
        EDL & 73.43 & 80.22 & 72.61 & 81.42 & 63.43 & 85.09\\
        \textbf{R2D2-Net}  & \textbf{98.75}  &  \textbf{98.84} & \textbf{99.64} & \textbf{99.65}& 99.31 & 99.69\\
        \bottomrule
        \end{tabular}
    }
    \label{tab: results_mnistOOD}
    \vspace{-4mm}
\end{table*}


\begin{table}[]
\caption{The OOD detection performance (\%) on DRD \citep{filos2019bdlb}, with
the LeNet \citep{lecun1989LeNet} architecture.
}
\centering
\scalebox{0.84}{
\begin{tabular}{lccc}
\toprule
\multicolumn{1}{l}{\textbf{Models}} &
\multicolumn{1}{c}{\textbf{AUROC}} &
\multicolumn{1}{c}{\textbf{AUPR}} \\ \hline
\multicolumn{1}{l}{DE} & 59.67   & 56.58 \\ 
\multicolumn{1}{l}{MCD \citep{gal2016mcdropout}} &  59.52  & 60.95  \\ 
\multicolumn{1}{l}{Gauss-SVI} & 69.12 & 74.80 \\
\multicolumn{1}{l}{Gauss-MFVI} & 63.56 & 66.79 \\ 
\multicolumn{1}{l}{Gauss-RAD} & 66.75 & 77.63 \\ 
\multicolumn{1}{l}{HS-SVI} & 69.80& 77.88\\
\multicolumn{1}{l}{SaS-SVI} & 70.64 & 75.19 \\
\multicolumn{1}{l}{DPN \citep{malinin2018DPN}} & 60.57 & 65.32 \\ 
\multicolumn{1}{l}{EDL \citep{sensoy2018EDL}} & 53.01   & 58.22  \\ 
\multicolumn{1}{l}{\textbf{R2D2-Net}} &  \textbf{71.04} & \textbf{78.11} \\ 
\bottomrule
\end{tabular}}
\label{tab: OOD Diabetes}
\end{table}

\begin{table}[h]
    \centering
    \caption{Summary of models used in the experiments and their depth.  Freq stands for frequentist networks, and Bayes stands for the Bayesian counterparts.}    
    \begin{tabular}{lcc}
    \toprule
    \textbf{Models} & \textbf{\# Params (Freq)} & \textbf{\# Params (Bayes)} \\ \hline
    LeNet    & 62K & 124K \\
    AlexNet  & 2.8M & 5.6M\\
    ResNet50     & 25.6M & 51.2M\\
    ResNet101 & 44.5M & 89M\\
    \bottomrule
    \end{tabular}
    \label{tab: architectures}
    \vspace{-2mm}
\end{table}

\subsection{Ablation Analysis}
\para{Performances with Various Architectures}
We further apply R2D2 layers to different neural network architectures to evaluate the performance.
We summarize the model architectures used in this experiment and their complexity in Table \ref{tab: architectures}.
We choose LeNet \cite{lecun1989LeNet} and AlexNet \cite{krizhevsky2012alexnet} to benchmark the performance of different BNN designs.
Table \ref{tab: ablation_architectures} presents the results on CIFAR 10. 
We observe that for most architectures our proposed BNN design achieves SOTA performance compared with existing BNN methods.
\begin{table*}[h]

\caption{
Image classification results of R2D2-Net with different model architectures compared to different combinations of priors and inference algorithms on the CIFAR 10 dataset.
\vspace{-2mm}
}
    \centering
    \scalebox{0.8}{
        \begin{tabular}{llcc|cc|cc|cccc p{1.3cm}}
        \toprule
        \multicolumn{2}{l}{\textbf{}} &
        \multicolumn{2}{c}{\textbf{LeNet}} &
        \multicolumn{2}{c}{\textbf{AlexNet}} &
        \multicolumn{2}{c}{\textbf{ResNet50}} &
        \multicolumn{2}{c}{\textbf{ResNet101}}
        \\
        \multicolumn{1}{l}{\textbf{Model/Prior}} &
        \multicolumn{1}{l}{\textbf{Inf. Alg. }} &
        \multicolumn{1}{c}{\textbf{AUROC}} & 
        \multicolumn{1}{c}{\textbf{ACC}} &
        \multicolumn{1}{c}{\textbf{AUROC}} & 
        \multicolumn{1}{c}{\textbf{ACC}} &
        \multicolumn{1}{c}{\textbf{AUROC}} & 
        \multicolumn{1}{c}{\textbf{ACC}} &
                \multicolumn{1}{c}{\textbf{AUROC}} & 
        \multicolumn{1}{c}{\textbf{ACC}} 
        \\ \hline
        Freq  &  &  91.24 & 61.21   & 92.70* & 65.03 & 96.23  &  79.25* & 96.75 & 79.20 \\ \hline
        DE &  & 93.75 & 63.11 & 90.06 & 62.94 & 96.60 & 77.62 & 96.82 & 79.01\\
        MCD &  & 91.50 & 58.76  &91.21   &  62.76& 96.44 & 77.24 & 96.83 & \textbf{79.54}* \\ \hline
        \multirow{5}{*}{Gauss} & SVI & 91.31& 60.03  & 91.21  & 62.64 & 95.59 & 73.62 & 95.53 & 73.34 \\
         & MFVI & 92.41 & 63.39 & 91.11& 59.27 & 96.48 & 78.19 & 95.65 & 73.37\\
        &Radial & 91.74 & 61.29 & 91.22 & 63.24 & 95.39 & 74.03 & 96.34 & 72.99 \\
         & SGMCMC & 80.34& 33.86 & 83.39 & 42.00 & 96.80 & 75.05 & 96.19 & 73.07 \\
         & SGLD & 81.78 & 36.47 & 89.34 & 52.22 & 88.08 & 48.00 & 88.09 & 49.34\\ \hline
        \multirow{3}{*}{SaS} & SVI & 91.87 & 60.50 & 91.69  & 60.84 & 94.18 & 74.15 & 94.17 & 70.26  \\
         & SGMCMC & 81.63 & 35.06 & 83.10 & 40.82 & 96.71 &78.91 & 96.76 & 79.36 \\
         & SGLD & 88.76 & 34.00 & 86.23  & 44.29 &  93.03 & 58.26 & 90.18 & 53.49 \\ \hline
        \multirow{2}{*}{HS} & SVI & \textbf{92.42}*  & 60.13 &  91.99 & 65.01 & 96.96  & 78.90 & 97.08 & 79.14\\
        & SGMCMC & 86.98 & 49.35 & 86.32 & 49.63  & 95.90 & 78.17 & 95.74 & 75.73\\ \hline
        \textbf{R2D2-Net} & SVGI & 90.43 & \textbf{61.53*} & \textbf{92.49} & \textbf{65.10}* & \textbf{96.97}* & \textbf{79.10} & \textbf{97.12}* & 79.20\\
        \bottomrule
        \end{tabular}}
    \label{tab: ablation_architectures}

\end{table*}
This demonstrates that the R2D2-Net performs satisfactorily on different architectures including modern architectures at the  large scale (e.g., ResNet101).

\section{Discussion}

We highlight the differences between our work and that of \citeauthor{zhang2020r2d2} \cite{zhang2020r2d2}.
Essentially, we adopt the prior in \citeauthor{zhang2020r2d2} and adapt the Gibbs inference algorithm to the deep learning context. 
The R2D2 prior has several  desirable properties: the highest concentration rate at zero and the heaviest tails, which are crucial to the development of BNN models. 
A well-known work is the Horseshoe BNN \cite{ghosh2019horseshoebnn} using the Horseshoe prior \cite{carvalho2009horseshoe}, which can effectively sparsify neural networks. We highlight the disadvantages of the Horseshoe prior (i.e., the lighter tail and lower concentration rate at zero) and show that the R2D2 prior is a better choice for variable shrinkage in neural networks.

\para{R2D2-Net for Sparsity-Induced Deep Learning. }
Sparsity-induced deep learning aims to resolve the excessive over-parameterization of modern deep neural networks \citep{sun2022consistentsparse, polson2018posteriorconcentr}. 
Bayesian methods impose sparsity on neural network weights via spike-and-slab priors, whereas the posterior contraction rate around the optimal predictor and the posterior consistency are the key factors to a good choice of prior.
The R2D2 prior has the near-minimax posterior contraction rate as the spike-and-slab priors.
It also yields a strongly consistent posterior \citep{zhang2020r2d2}, which makes it a competitive candidate over the existing spike-and-slab priors for sparsity-induced deep learning.

The R2D2 prior is also able to facilitate neural network sparsification, which aims to prune the neural network for smaller time complexity.
Existing works widely adopt the variational dropout or Horseshoe prior to shrink unnecessary neurons to achieve neural network sparsification.
However, the relatively low concentration rates around zero of these priors make the pruning process ineffective, while the light tails of these priors cause some important features to be overshrunk. 
Since the R2D2 prior has a relatively higher concentration rate around zero and a heavier tail compared to existing sparsity-induced priors, it also performs superiorly in neural network sparsification. 

\para{The Analogy of $R^2$. } The R2D2 prior is loosely based on a prior on the goodness-of-fit in a regression model. Under the deep learning settings, the analogy of $R^2$ may not be applicable in the settings of DNN/BNN, as it is difficult to interpret the ``goodness-of-fit''. However, the properties of the R2D2 prior (i.e., the high spiking rate and heavy tail) still hold as we place such a prior on each individual neuron. 

\para{Different Shrinkage Profiles. } Recently, many new shrinkage priors have been proposed by introducing more parameters in the prior to more flexibly determine the shrinkage behaviours. 
For instance, the triple gamma prior \cite{cadonna2020triple} added an extra layer to the double gamma prior to more specifically capture the shrinkage patterns. 
It possesses the concentration rate of $\mathcal{O}\big(( {1}/{\sqrt{\beta}})^{1 - 2 a^\zeta}\big)$ at zero, which is comparatively lower than that of the R2D2 prior.
And it possesses  a tail thickness of $ \mathcal{O}\big(( {1}/{\sqrt{\beta}})^{2 c^\zeta + 1}\big)$ which is comparatively thinner than that of the R2D2 prior.
In modern deep learning settings, effective shrinkage and feature preservation are important for optimal DNN and BNN performance, hence the R2D2 prior remains the competitive candidate for priors on neural network weights.

\para{Hyperparameters. } We adopt grid search to perform hyperparameter tuning. However, \citeauthor{gruber2022forecasting} \cite{gruber2022forecasting} proposed a data-driven approach such that the key hyperparameters (e.g., shrinkage) can be estimated from data. Hence, including the parameters into the stochastic back-propagation may be feasible in future extensions of the R2D2-Net.

\para{Limitations}
Our proposed method tackles the scalability constraints of Bayesian neural networks and is validated on some modern architectures (e.g., ResNet-based). 
%
%
Due to a lack of mature research in Bayesian designs of more modern architectures (such as Bayesian attention mechanisms and transformers), the extension of R2D2-Net to these architectures would be non-trivial, although it opens the possibilities of Bayesian foundation models.
On the other hand, we take Monte Carlo samples of weights from the posterior distribution, which could potentially be a computational burden.
Integration of recent efficient sampling techniques of BNNs \cite{dusenberry2020rank1BNN, franchi2023LPBNN} would decrease the posterior inference complexity.

Moreover, the Gibbs sampling based on conditional distributions may not tackle the multimodal posteriors satisfactorily. With the recent development of multimodal Gibbs inference \cite{chen2024diffusiveGibbs}, it would be interesting to be integrated into the SVI paradigm in future works.

\section{Conclusion}
In this work, we propose a novel BNN design --- the R2D2-Net. 
We develop a variational Gibbs inference algorithm to better approximate the posterior distributions of
the network weights. 
Extensive experiments on synthetic and real datasets validate the performance of our proposed BNN design on both image classification and image uncertainty estimation tasks. 
Our proposed method can be potentially applied to different data domains, such as graphs and Bayesian graph neural networks. 
The R2D2-Net also has great real application potential in reinforcement learning, recommendation systems, and biomedical imaging for its capability in predictive inference and uncertainty estimation. 

\para{Acknowledgement.} 
We thank the Associate Editor and referees for many constructive suggestions that have significantly improved the paper. This work was supported in part by the Research Grants Council of Hong Kong (17308321, 27206123, C5055-24G, and T45-401/22-N), Patrick SC Poon endowment fund, Hong Kong Innovation and Technology Fund (ITS/273/22 and ITS/274/22), National Natural Science Foundation of China (No. 62201483), and Guangdong Natural Science Fund (No. 2024A1515011875).

\bibliographystyle{plainnat}
\bibliography{references}

\vspace{3mm}

\begin{wrapfigure}{l}{0.2\textwidth}
    \centering
    \includegraphics[width=0.18\textwidth, height=0.17\textwidth]{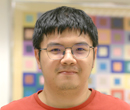}
    \vspace{-1mm}
\end{wrapfigure}
\textbf{Tsai Hor Chan} received the BSc degree in Actuarial science from University of Hong Kong in 2020. He is currently a final-year PhD student in Department of Statistics and Actuarial Science, University of Hong Kong. His research interest focuses in Bayesian nonparametrics, statistical deep learning, medical informatics, and computer vision.

\vspace{3mm}

\begin{wrapfigure}{l}{0.2\textwidth}
    \centering
    \includegraphics[width=0.16\textwidth, height=0.19\textwidth]{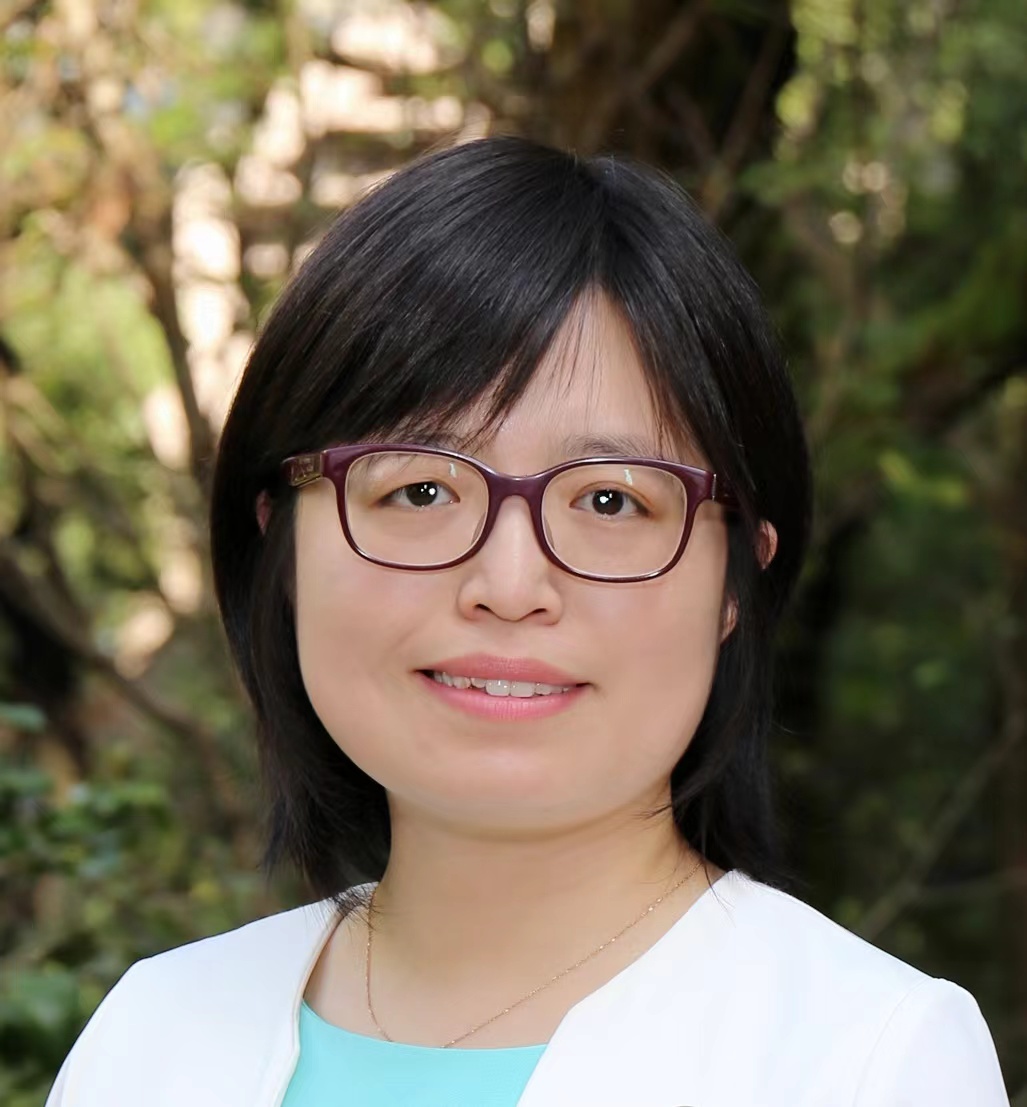}
    \vspace{-1mm}
\end{wrapfigure}
\textbf{Yan Dora Zhang} received  Ph.D. in Statistics from North Carolina State University in 2016. She is currently an Assistant Professor in Department of Statistics and Actuarial Science, University of Hong Kong. Her main research areas include statistical genetics, biostatistics, bioinformatics, health informatics, and public health.

\vspace{3mm}

\begin{wrapfigure}{l}{0.2\textwidth}
    \centering
    \includegraphics[width=0.17\textwidth, height=0.18\textwidth]{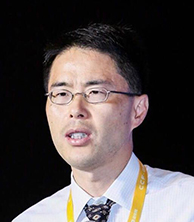}
    \vspace{-1mm}
\end{wrapfigure}
\textbf{Guosheng Yin} (Senior Member, IEEE; Fellow, American Statistical Association; Fellow, Institute of Mathematical Statistics) received Ph.D. in Biostatistics from University of North Carolina at Chapel Hill in 2003.  In 2003-2009, he worked as Assistant Professor and also Associate Professor (tenured) in Department of Biostatistics at University of Texas M.D. Anderson Cancer Center. He worked as Head of Department of Statistics and Actuarial Science at University of Hong Kong in 2017-2023 and then as Chair in Statistics at Imperial College London in 2023-2024. He is currently Chair Professor of Statistics and Patrick SC Poon Endowed Professor in Department of Statistics and Actuarial Science at University of Hong Kong.  His main research areas include AI, machine learning, clinical trial methodology, adaptive design, Bayesian methods, and survival analysis. He has published over 260 peer-reviewed papers and two books in the areas of clinical trial design and methods.

\vspace{3mm}

\begin{wrapfigure}{l}{0.2\textwidth}
    \centering
    \includegraphics[width=0.16\textwidth, height=0.19\textwidth]{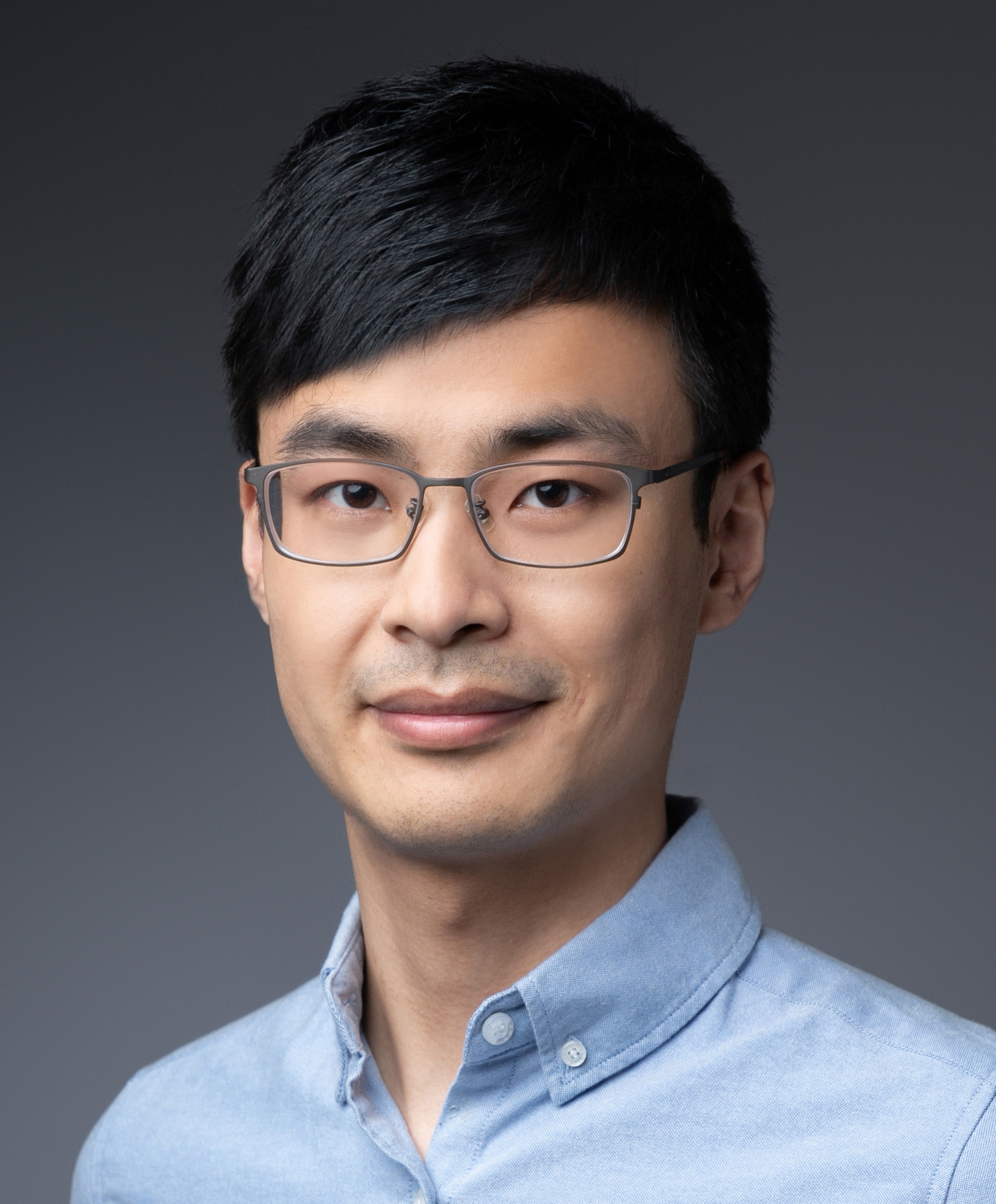}
    \vspace{-1mm}
\end{wrapfigure}

\textbf{Lequan Yu} (Member, IEEE) received the BEng degree from Department of Computer Science and Technology, Zhejiang University, Hangzhou, China, in 2015, and the PhD degree from Department of Computer Science and Engineering, Chinese University of Hong Kong, Hong Kong, in 2019. He conducted his postdoctoral training at Stanford University during 2019--2021. He is currently an Assistant Professor in Department of Statistics and Actuarial Science, University of Hong Kong. His research focuses on medical image analysis, computer vision, machine learning and AI in healthcare.

\appendix
\onecolumn
\para{Overview. } In the supplementary material, we first discuss the technical details in Section  \ref{sec: technical details}. 
Section \ref{sec: experiments} presents additional experiment results of the R2D2-Net and more details on the implementation and hyperparameters.
We provide further details on composing network architectures with R2D2 layers and illustrate architectural samples of the R2D2-Net in Section \ref{sec: archi}.
The properties of distributions used in our work are given in Section \ref{sec: distributions}, and derivations of the KL divergences presented in the main text are detailed in Section \ref{sec: kl}. 
Information on other global--local shrinkage priors is presented in Section \ref{sec: gl-priors}.
%
%
%
Finally, we provide the definitions of  uncertainty measures in Section \ref{sec: uncertainty-measures} and evaluation metrics used in the experiments in Section \ref{sec: ev-metrics}. 



%
%

\section{Technical Details} \label{sec: technical details}
The proof of Theorem 1
is based on \cite{sun2022spikeandslab}, which formulates the convergence rates of sparse DNNs.
%
We consider a general prior setting that all entries of $\mathbf{w}$ are subject to independent continuous prior $\pi_b$, i.e., $\pi(\mathbf{w}) = \prod_{j=1}^{K_n}\pi_b(w_j)$.

\begin{theorem} \label{lemma: general convergence rate}
    (Convergence rate under the general prior \citep{sun2022sparseDNNTheories,sun2022spikeandslab}) For deep forward neural networks with $L$ hidden layers, suppose the regularity conditions A.1--A.3 in the main text hold. Let $\epsilon_n \in (0, 1]$ be a sequence such that $n\epsilon_n \to \infty$ and $\epsilon_n > M \varpi_n$ for some large constant $M$.  If for some $\tau > 0$ the prior distribution satisfies that 
    \begin{align}
        \log(1 / \underline{\pi}_b) = O(H_n \log n + \log & \bar{L} )\label{cond: 1}\\ 
        \pi_b\{[-\eta_n, \eta_n]\} \geq 1 - \dfrac{1}{K_n} e^{ - \tau (H_n \log n + \log \bar{L} + \log D_n)} \text{ and } & \pi_b\{[-\eta_n', \eta_n']\} \geq 1 - \dfrac{1}{K_n} \label{cond: 2}\\ 
        - \log [K_n \pi_b(|w_j| > M_n)] \prec n \epsilon_n^2,  \label{cond: 3}
    \end{align}
    where $\eta_n < 1/ \{ \sqrt{n}K_n (n/H_n)^{H_n}(c_0 M_n)^{H_n}\}, $ $\eta_n' < 1/\{\sqrt{n}K_n (r_n/H_n)^{H_n}(c_0E_n)^{H_n}\}$ with some $c_0 > 1$, $\underline{\pi}_b$ is the minimal density value of $\pi_b$ within interval $[-E_n - 1, E_n + 1]$, and $M_n$ is some sequence satisfying $\log(M_n) = O(\log(n))$. Then, there exists a sequence $\epsilon_n$, satisfying $n\epsilon_n^2 \asymp r_n H_n \log n + r_n \log \bar{L}  + s_n \log D_n + n \varpi_n^2$, and $\epsilon_n \succ 1$, such that  
     the following results hold:
    \begin{enumerate}
        \item For all sufficiently large $n$, $\mathbb{P}^*\left\{ p[d(p_{\mathbf{w}}, p_{\bm \mu^*}) > 4 \epsilon_n | \mathcal{D}] \geq 2 e^{-n \epsilon_n^2 / 4} \right\} \leq 2 e^{-n \epsilon_n^2 / 4}$
        \item For all sufficiently large $n$, $\mathbb{E}^*_{\mathcal{D}} \left\{ p[d(p_{\mathbf{w}}, p_{\bm \mu^*}) > 4 \epsilon_n | \mathcal{D}]\right\} \leq 4 e^{- n \epsilon_n^2 / 2}$,
    \end{enumerate}
    where  $\mathbb{P}^*$ and $\mathbb{E}^*$ denote the respective probability measure and expectation with respect to the data $\mathcal{D}$, and $p[\cdot]$ represents the posterior distribution.
\end{theorem}

\begin{lemma} \label{lemma: r2d2 density}
    (Density of the R2D2 prior \cite{zhang2020r2d2}) The density of the R2D2 prior, denoted as $\pi_{\text{R2D2}}(\beta)$, has the form,
    \begin{align} \label{eq: R2D2 Density}
        \pi_{\text{R2D2}}(\beta) =  \dfrac{2^{a_\pi \Gamma(a_\pi + b)}}{\Gamma(a_\pi)\Gamma(b)} \int_0^{\infty} \exp (-|\beta|x) \dfrac{x^{2b}}{(x^2 + 2)^{a_\pi + b}} dx.
    \end{align}
    Then as $\beta \to \infty$, we have 
    \begin{align*}
        \pi_{\text{R2D2}}(\beta) = \mathcal{O} \left( \dfrac{1}{\beta^{2b+1}}\right).
    \end{align*}
\end{lemma}

\begin{lemma} \label{lemma: g_beta}
    For $u > 0$, 
    $k_n \asymp \sqrt{(r_n\log D_n) / n} / D_n$, let $g(\beta) = \pi_{\text{R2D2}}(\beta) $, 
    and then we have 
    \begin{align*}
        1 - \int_{-k_n}^{k_n} g(\beta) d\beta = C_1^*U_1(k_n^2) + C_2^*U_2(k_n^2) + C_3^*U_3(k_n^2),
    \end{align*}
    where 
    \begin{align*}
        C_1^* &= \dfrac{1}{\sqrt{\pi}\Gamma(a_\pi)\Gamma(b)} \Gamma(-a_\pi)\Gamma\left( \dfrac{1}{2} - a_\pi\right) \Gamma\left(  a_\pi + \dfrac{1}{2} \right)\Gamma(1+a_\pi) < 0,\\
         C_2^* &= \dfrac{1}{\sqrt{\pi}\Gamma(a_\pi)\Gamma(b)} \Gamma(a_\pi)\Gamma\left(\dfrac{1}{2} \right) \Gamma\left(\dfrac{1}{2} \right)\Gamma(1-a_\pi) > 0,\\
         C_3^* &= \dfrac{1}{\sqrt{\pi}\Gamma(a_\pi)\Gamma(b)} \Gamma\left(  a_\pi - \dfrac{1}{2} \right) \Gamma\left(- \dfrac{1}{2} \right) \Gamma\left(\dfrac{3}{2} \right) \Gamma\left(\dfrac{3}{2} - a_\pi\right) > 0,\\
         U_1(k_n^2) &= \sum_{j=0}^{\infty} (-1)^j u_1(j, k_n^2) ,\\
         u_1(j, k_n^2) &= \dfrac{\Gamma(a_\pi + b +j)}{\Gamma(1 + a_\pi + j)\Gamma(\frac{1}{2} + a_\pi + j)} \dfrac{\left(\frac{k_n^2}{2} \right)^{j + a_\pi}}{j!}, \\
          U_2(k_n^2) &= \sum_{j=0}^{\infty} (-1)^j u_2(j, k_n^2), \\
         u_2(j, k_n^2) &= \dfrac{\Gamma( b +j)}{\Gamma(1 - a_\pi + j)\Gamma(\frac{1}{2} + j)} \dfrac{\left(\frac{k_n^2}{2} \right)^{j}}{j!},\\
          U_3(k_n^2) &= \sum_{j=0}^{\infty} (-1)^j u_3(j, k_n^2), \\
         u_3(j, k_n^2) &= \dfrac{\Gamma(\frac{1}{2}+ b +j)}{\Gamma(\frac{3}{2} - a_\pi + j)\Gamma(\frac{3}{2} + j)} \dfrac{\left(\frac{k_n^2}{2} \right)^{j + \frac{1}{2}}}{j!}.
    \end{align*}
\end{lemma}

\para{Proof of Theorem 1. } It suffices to verify the conditions listed in Theorem \ref{lemma: general convergence rate} with $M_n = \max \{ \sqrt{2n}b_n, E_n\}$ .

Condition \ref{cond: 1} can be verified via the density of the R2D2 prior. We consider the tail of the R2D2 prior since its density has the minimum values at its tails, which is specified by Lemma \ref{lemma: r2d2 density} and $E_n / \{ L_n \log n + \log \bar{H} \}^{1/2} \lesssim b \lesssim n^\alpha$.
Condition \ref{cond: 2} can be verified by Lemma \ref{lemma: g_beta} by substituting $\eta_n$ and $\eta_n'$ into $k_n$, and note that 
\begin{align*}
    1 - \int_{-k_n}^{k_n} g(\beta) d\beta &\leq C_1^* \left\{u_1(0, k_n^2) - u_1(1, k_n^2)\right\} + C_2^* u_2(0, k_n^2) + C_3^* u_3(0, k_n^2) \\
    & = 1 - k_n^{2a_\pi} \left\{ - \dfrac{\Gamma(-a_\pi)\Gamma(\frac{1}{2} - a_\pi)\Gamma(a_\pi + b)}{\sqrt{\pi}\Gamma(a_\pi)\Gamma(b)2^{a_\pi}} - C_4^*k_n^2 - C_5^* k_n^{1 - 2 a_\pi}\right\}\\
    & = 1 - k_n^{2a_\pi} \left\{ - \dfrac{\Gamma(1-a_\pi)\Gamma(\frac{1}{2} - a_\pi)\Gamma(a_\pi + b)}{\sqrt{\pi}\Gamma(b)2^{a_\pi}} - C_4^*k_n^2 - C_5^* k_n^{1 - 2 a_\pi}\right\}\\
    & \to 1 - k_n^{a_\pi} \leq D_n^{- (1+u)},
\end{align*}
where $k_n \to 0$ and $a_\pi \leq \dfrac{\log (1 - D_n^{-(1+u)})}{2 \log k_n} \to 0$, and $C_4^*, C_5^* \geq 0$.
Condition \ref{cond: 3} can be verified through Lemma \ref{lemma: g_beta} with $E_n / \{ L_n \log n + \log \bar{H} \}^{1/2} \lesssim b \lesssim n^\alpha$.

\noindent







\section{Further Discussions}


\subsection{R2D2-Graph Neural Nerworks}
The R2D2-net can potentially provide a good solution to several challenging problems in graph neural networks (GNNs).
GNNs have been a powerful tool in many prediction tasks \citep{bianchi2021armaGNN,chen2018fastgcn, chen2021MLRGCN,gao2021topologyGPN,  velivckovic2017GAT,kipf2016gcn,xu2018GIN,zheng2020neuralsparse}.
However, the widely adopted message-passing mechanism suffers from the unnecessary (i.e., task-irrelevant) features and hinders the predictive and generalization performance. 
The goal of graph sparsification is to find the smallest possible subgraph that preserves the properties of the original graph.
Several sparsity-induced methods \citep{Bin2021GraphmaxFT, wang2023sparseEEG,zheng2020neuralsparse}  have been introduced to sparsify the graph data for robust learning.
Introducing the R2D2-net to GNN can potentially impose the best shrinkage effects on unnecessary edges or nodes (due to its high concentration rate at zero) and preserve the most meaningful subgraphs (due to the heavy tail property).
The shrinkage property also allows for controlling the receptive fields of the nodes (i.e., only keeping the most relevant features to the node), addressing the well-known over-smoothing problem in GNNs. 
This would enable the development of deeper GNNs.
Hence, designing R2D2-GNN is a promising future direction to facilitate robust graph representation learning, and it can potentially be scalable to deep GNNs.

\section{Additional Information on Experiments} \label{sec: experiments}

We provide additional information on baselines and experiment results in this section, including additional settings of hyperparameters and implementation. 

\subsection{Additional Experiment Results}
\para{Additional Image Classification Results.}
We present additional results on image classification and ablation studies with different architectures using more evaluation metrics. 
Table \ref{tab: CIFAR_Le} shows the image classification results on CIFAR 10 with the architecture fixed as LeNet. 
We observe that the improvement of R2D2-Net is less significant compared to that using AlexNet. 

\begin{table*}[h]
\caption{Image classification results of our proposed method on CIFAR 10 and CIFAR 100 with the LeNet \citep{lecun1989LeNet}. 
%
}
    \centering
        \begin{tabular}{lcccccc p{1.3cm}}
        \toprule
        \multicolumn{1}{l}{\textbf{}} &
        \multicolumn{3}{c}{\textbf{CIFAR 10}} &
        \multicolumn{3}{c}{\textbf{CIFAR 100}}
        \\
        \multicolumn{1}{l}{\textbf{Model}} &
        \multicolumn{1}{c}{\textbf{AUROC}} & 
        \multicolumn{1}{c}{\textbf{Accuracy}} &
        \multicolumn{1}{c}{\textbf{Macro-F1}} &
        \multicolumn{1}{c}{\textbf{AUROC}} & 
        \multicolumn{1}{c}{\textbf{Accuracy}} &
        \multicolumn{1}{c}{\textbf{Macro-F1}} 
        \\ \hline
        Frequentist  & 91.38 & 62.24 & 62.34 & 89.45  & 30.51  & 29.64  \\ \hline
        Gaussian BNN  & 91.31  & 60.03  & 59.55  & 89.17 & 25.79 & 25.08 \\
        MC Dropout & 91.50  & 58.76  &  59.4  &  90.65 & 27.23 & 25.83\\ 
        MFVI & 92.22 & 61.94 & 61.71 & 88.90 & 29.63 & 29.06\\ 
        Radial BNN & 92.13 & 61.71 & 61.30 & 89.77 & 30.27 & 29.8\\
        Deep Ensembles &  92.74 & 64.26 & 64.14 & 89.37 & 30.04 & 29.44 \\
        Horseshoe BNN & 92.42 & 60.13 &  59.80 & 85.88 &  17.94 & 16.01\\
        \textbf{R2D2-Net}  & 92.39 & 62.14 & 62.02 & 88.59 & 30.51 & 29.82 \\
        \bottomrule
        \end{tabular}
    \label{tab: CIFAR_Le}
\end{table*}

\begin{table*}[t]
\caption{Image classification results of our proposed method on CIFAR 10 and CIFAR 100 with the AlexNet \citep{krizhevsky2012alexnet}. 
Standard deviations are given in brackets. 
%
\textbf{Boldface} represents the best performance among BNN designs, while * represents the best performance among all models.}
\setlength{\tabcolsep}{2mm}
    \centering
    \scalebox{0.9}{
        \begin{tabular}{lccc|ccc p{1.3cm}}
        \toprule
        \multicolumn{1}{l}{\textbf{}} &
        \multicolumn{3}{c}{\textbf{CIFAR 10}} &
        \multicolumn{3}{c}{\textbf{CIFAR 100}}
        \\
        \multicolumn{1}{l}{\textbf{Model}} &
        \multicolumn{1}{c}{\textbf{AUROC}} & 
        \multicolumn{1}{c}{\textbf{Accuracy}} &
        \multicolumn{1}{c}{\textbf{Macro-F1}} &
        \multicolumn{1}{c}{\textbf{AUROC}} & 
        \multicolumn{1}{c}{\textbf{Accuracy}} &
        \multicolumn{1}{c}{\textbf{Macro-F1}} 
        \\ \hline
        Frequentist NN & 92.70 (1.5)* & 65.03 (1.4) & 64.9 (1.6)* & 90.95 (0.2) & 31.05 (0.4)& 31.21 (0.6)  \\ \hline
        Gaussian BNN & 91.37 (1.2) & 60.28 (1.5) & 60.78 (1.4)  & 87.24 (1.2) & 23.60 (0.6) & 22.02 (0.8) \\
        MC Dropout & 90.09 (0.2) & 55.08 (0.6) &  54.22 (0.3) & 87.67 (1.3)& 21.92 (1.1)  & 19.74 (1.1) \\ 
        MFVI &91.11 (0.9) & 59.27 (1.1) & 61.82 (0.5) & 87.69 (0.9) & 23.06 (0.2) & 22.34 (0.2)\\
        Radial BNN & 91.22 (0.8) & 63.24 (0.8) & 62.48 (0.9) & 89.20 (1.0) & 25.70 (0.5) & 24.89 (0.7)\\
        Deep Ensembles & 90.67 (0.6) & 62.41 (0.5) & 62.43 (0.4) & 87.97 (0.9) & 24.63 (0.4) & 23.94 (0.5) \\ 
        Horseshoe BNN & 91.99 (0.8) & 65.01 (0.3) &  64.7 (0.3) & 91.37 (0.2) &  33.27 (0.3) & 34.02 (0.3)\\
        \textbf{R2D2-Net}  & \textbf{92.49 (0.2)} & \textbf{65.10 (0.02)*} & \textbf{65.14 (0.06)} & \textbf{91.41 (0.03)}* & \textbf{36.12 (0.5)}* & \textbf{34.83 (0.4)}* \\
        \bottomrule
        \end{tabular}}
    \label{tab: CIFAR_Alex_Full}
\end{table*}

\para{Implementation Details and Hyperparameters}
The proposed method is implemented in Python with \textit{Pytorch} library on a server equipped with four NVIDIA TESLA V100 GPUs. 
All methods are trained for 1000 epochs for image classification and 100 epochs for OOD detection with possible early stopping.
We randomly initialize the weights of each architecture (i.e., train from scratch).
We select the checkpoint with the largest validation AUROC as the testing checkpoint. We use \textit{Adam} as the optimizer with a learning rate of 0.0005 and a weight decay of 0.0005. The batch size is 1024. The dropout ratio is 0.2 for MC Dropout \citep{gal2016mcdropout}.
We set a universal annealing rate of 0.001 for the KL loss since we did not encounter any KL vanishing problem. 
Data augmentation procedures such as colour jittering and random cropping and flipping are applied to regularize the learning process.

\begin{figure}[h]
        \centering
        \includegraphics[width=0.95\textwidth]{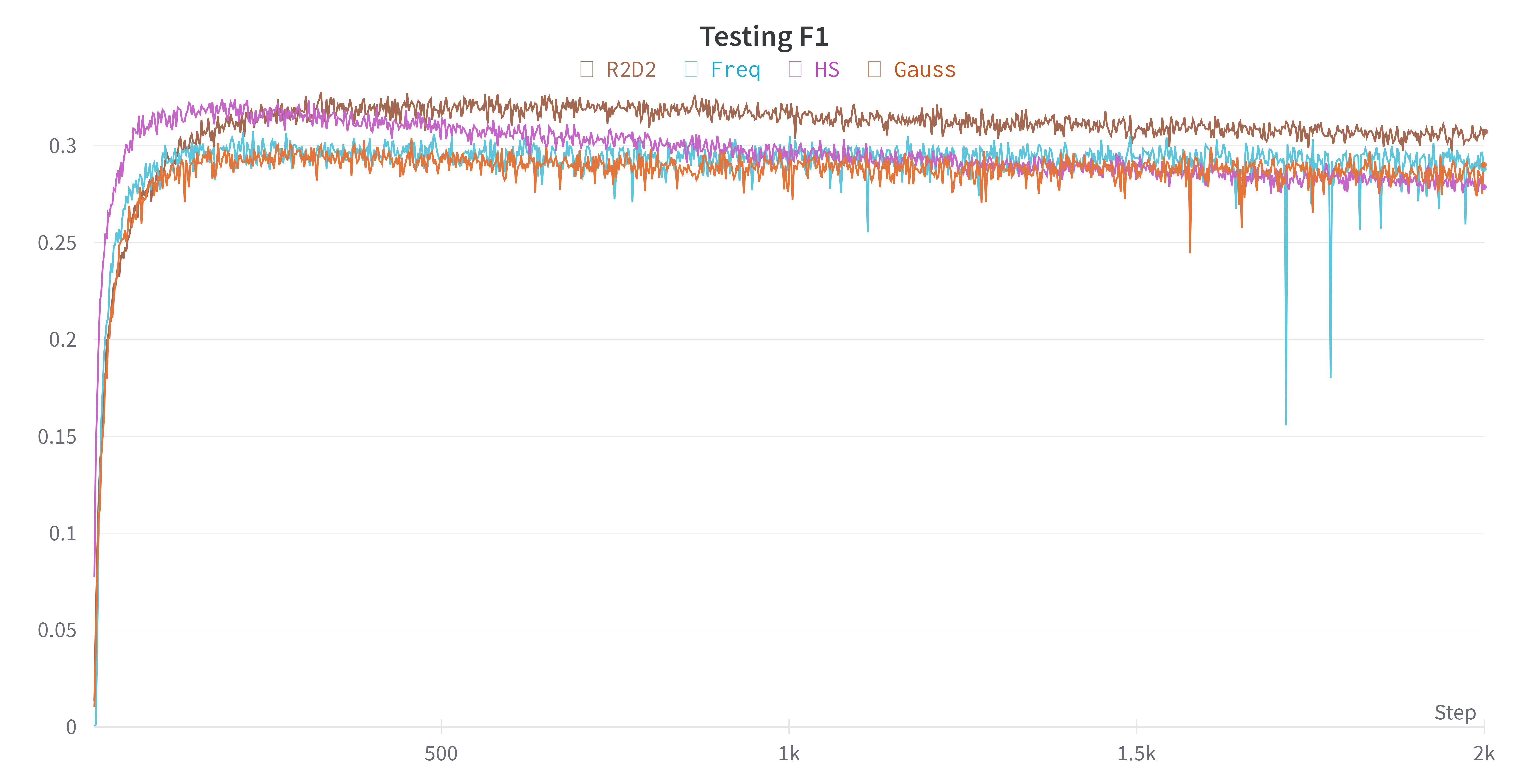}
        \caption{Comparison of training curves.}
        \label{fig: training_curve}
\end{figure}

\para{Learning Curve Comparison}
We compare the learning curves from different BNN designs in Figure \ref{fig: training_curve}.
We observe that the R2D2 Net has consistent higher testing performance from epoch to epoch. 

\subsection{Additional Details on Model Architectures} \label{sec: archi}

\para{Compose Network Architecture with R2D2 Layers.}
With the given marginal weight distributions in Eq. (1) from the main text, we can construct the layers of the R2D2-Net.
Specifically, we consider two basic operations in a neural network --- the linear layer and the convolutional layer. 
Let $\mathbf{w}_l$ be the vector of all weight parameters of the $l$-th layer. 
The distribution of the $j$-th element $w_{jl}$ follows the R2D2 distribution given in Eq. (1) in the main manuscript.
We compose the neural network architecture by specifying a combination of convolutional layers and linear layers. 
Figure 2 in the main manuscript presents the conditional dependencies of the R2D2 design and the training paradigm. 
As an illustrative example, the visualization of R2D2 LeNet is provided in the appendix.
Each linear layer and convolutional layer are replaced by the R2D2 counterparts (i.e., the R2D2 linear and R2D2 conv layers), while the pooling layers and activation layers remain the same as in their frequentist designs. 

\begin{figure*}[h]
        \centering
        \includegraphics[width=0.9\textwidth]{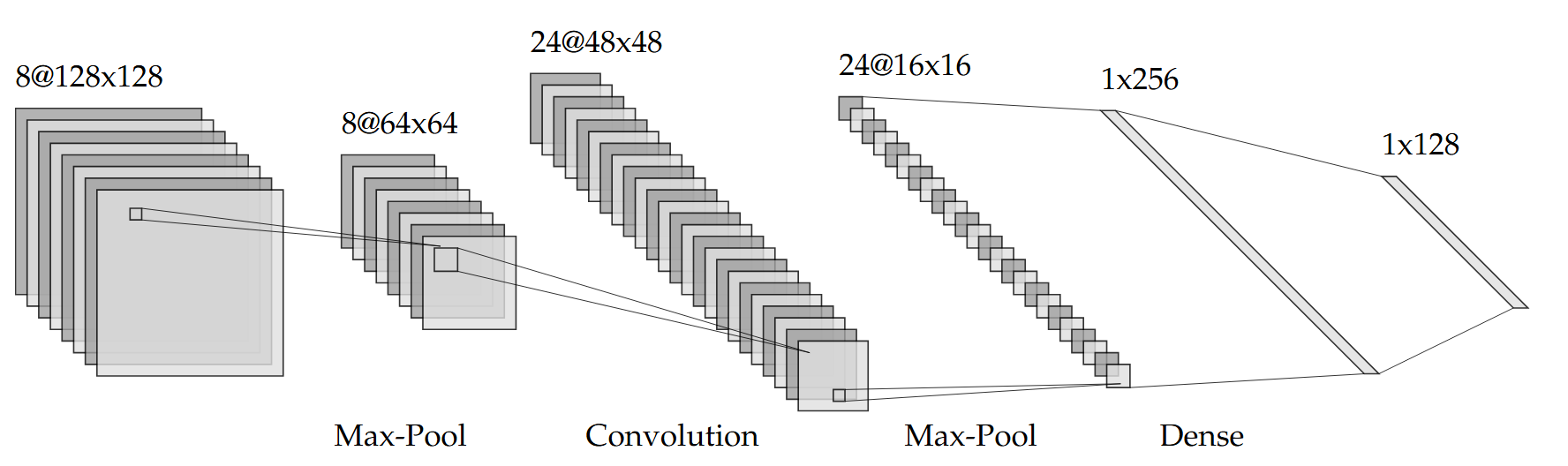}
        \caption{Example of the R2D2 LeNet architecture. Each convolutional or linear layer is replaced by its R2D2 design (i.e., R2D2 Linear or R2D2 Conv).}
        \label{fig:r2d2_lenet}
\end{figure*}

\para{Summary of Model Architectures and Complexity.}
Figure \ref{fig:r2d2_lenet} presents an example of the LeNet \citep{lecun1989LeNet} architecture composed by R2D2 layers, where each convolutional layer and each linear layer are replaced by its corresponding BNN design (e.g., R2D2 linear layer or R2D2 conv layer).

\subsection{Hyperparameter Settings}
The hyperparameter settings for different priors are given as follows:
\begin{itemize}
\item Number of posterior samples (during inference): 100
    \item Gaussian BNN:
    \begin{itemize}
        \item $\rho_0 \sim \mathcal{N}(-3, 0.1^2)$
    \end{itemize}
    \item Horseshoe
    \begin{itemize}
        \item Global shrinkage $b_g$ = 1.0
        \item Local shrinkage $b_0$ = 1.0
        \item $\rho_0 \sim \mathcal{N}(-3, 0.1^2)$
    \end{itemize}
    \item R2D2
    \begin{itemize}
    \item $a_\pi = 0.6$
    \item $b = 0.5$
    \item prior mean of $w_{jl} =  0$
    \item $\rho_0 \sim \mathcal{N}(-3, 0.1^2)$ 
    \end{itemize}
\end{itemize}

\section{Distributions} \label{sec: distributions}


\para{Expectations of Generalized Inverse Gaussian.}
We provide the well-known results of the expectations of functions of  $X \sim \text{giG}(\chi, \rho, \lambda_0)$ for completeness:
\begin{align*}
    \mathbb{E}(X) &= \dfrac{\sqrt{\chi}K_{\lambda_0+1}(\sqrt{\rho\chi})}{\sqrt{\lambda_0}K_{\lambda_0+1}(\sqrt{\rho\chi})},\\
    \mathbb{E}\left(\dfrac{1}{X}\right) & = \dfrac{\sqrt{\rho}K_{\lambda_0+1}(\sqrt{\rho\chi})}{\sqrt{\chi}K_{\lambda_0+1}(\sqrt{\rho\chi})} - \dfrac{2\lambda_0}{\chi},\\
    \mathbb{E}(\log X) & = \log \dfrac{\sqrt{\chi}}{\sqrt{\rho}} + \pdv*{\log K_{\lambda_0}(\sqrt{\rho\chi})}{\lambda_0}.
\end{align*}
The derivative term in the above equation does not have an analytical form and therefore needs to be computed numerically. 

\section{Kullback–Leibler (KL) Divergence} \label{sec: kl}

We provide detailed derivations of the KL divergences introduced in the main text. 

\para{KL Divergence of Gamma Distributions.}
Define the integral 
\begin{align*}
  I(a,b,c,d)  = \int_0^\infty \log \left( \dfrac{e^{x/a}x^{b-1}}{a^b\Gamma(b)}\right)
     \dfrac{e^{x/c}x^{d-1}}{c^d\Gamma(d)}dx,  
\end{align*} and then we have
\begin{equation} \label{eq: I}
     I(a,b,c,d) =  -\dfrac{cd}{a} - 
    \log (a^b \Gamma(b))  + (b-1)\psi(d) + (b - 1) \log(c)\\,
\end{equation}
where $\psi$ is the digamma function. 
The KL divergence between two Gamma distributions can be obtained in a closed form as 
\begin{align*}
    {\rm KL}(\text{Ga}(a, b) \Vert \text{Ga}(c, d)) =  I(a,b,c,d) - I(c,d,c,d).
\end{align*}

\para{KL Divergence of Multivariate Normal Distributions.}
The KL divergence of two multivariate normal distributions $\mathcal{N}(\bm \mu_1, \bm \Sigma_1)$ and $\mathcal{N}(\bm \mu_2, \bm \Sigma_2)$ is 
\begin{align*}
    {\rm KL} (\mathcal{N}( \bm \mu_1, \bm \Sigma_1) 
\Vert \mathcal{N}(\bm \mu_2, \bm \Sigma_2)  = 
\dfrac{1}{2} \Big[\log \dfrac{|\bm \Sigma_2|}{|\bm \Sigma_1|} - p  
 + {\rm tr}\{\bm \Sigma_2^{-1}\bm \Sigma_1\} + (\bm \mu_2 - \bm\mu_1)^\top \bm \Sigma_2^{-1}(\bm \mu_2 - \bm \mu_1) \Big].
\end{align*}

\para{KL Divergence of Shrinkage Parameters.} 
The closed form of ${\rm KL}(q(\xi | \cdot) \Vert \pi(\xi))$ is given by
\begin{align*}
    &{\rm KL}(q(\xi | \cdot) \Vert \pi(\xi)) \\
    =&
    \mathbb{E}_{q(\xi|\cdot)}[\log q(\xi|\cdot) - \log \pi(\xi)]\\
     =&\mathbb{E}_{q}\left[\log \left(\dfrac{(1 + \omega_l)^{a_l + b_l}}{\Gamma(a_l + b_l)}\xi_l^{a_l + b_l - 1}e^{-(1 + \omega_l) \xi_l}\right)\right] - \mathbb{E}_{q}\left[\log \left(\dfrac{1}{\Gamma(b_l)}\xi_l^{b_l - 1}e^{- \xi_l}\right)\right] \\
    =& I(a_l + b_l, 1 + \omega_l, 1, b_l) - I(1, b_l, 1, b_l),
\end{align*}
where the integral $I$ is defined in Eq. (\ref{eq: I}).

The closed form of ${\rm KL}(q(\omega_l | \cdot) \Vert \pi(\omega_l))$ is given by
\begin{align*}
    &{\rm KL}(q(\omega_l | \cdot) \Vert \pi(\omega_l)) \\
    =& \mathbb{E}_{q(\omega_l|\cdot)}[\log q(\omega_l|\cdot) - \log \pi(\omega|\xi_l)]\\
     =& \mathbb{E}_{q}\left[\log \left(\dfrac{(\rho/\chi)^{\lambda_0 /2}}{2K_{\lambda_0}(\sqrt{\rho\chi})}\omega_l^{\lambda_0-1}e^{(-\rho\omega_l+\chi/\omega_l)/2}\right)\right]  - \mathbb{E}_{q}\left[\log \left(\dfrac{\xi_l^{a_l}}{\Gamma(a_l)}\omega_l^{a_l - 1}e^{-\omega_l\xi_l}\right)\right]\\
     =& \dfrac{\lambda_0}{2}\log \dfrac{\rho}{\chi} - \log 2 - \log K_{\lambda_0}(\sqrt{\rho\chi}) + (\lambda_0 - 1)\mathbb{E}_{q}[\log \omega_l] - \dfrac{1}{2}\mathbb{E}_{q}(\rho \omega_l + \dfrac{\chi}{\omega_l}) \\
    & - \rho \log \xi + \log \Gamma(a_l) - (a_l - 1) \mathbb{E}_{q}[\log \omega_l] + \xi \omega_l.
\end{align*}

The closed form of the KL divergence of $\psi_{jl}$ is given by
\begin{align*}
    &{\rm KL}(q( \psi_{jl} | \cdot) \Vert \pi(\psi_{jl})) \\
    =& \mathbb{E}_{q(\psi_{jl}|\cdot)}[\log q(\psi_{jl}|\cdot) - \log \pi(\psi_{jl})]\\
     =& \mathbb{E}_{q(\psi_{jl}|\cdot)}\left[\log \left(\dfrac{1}{\psi_{jl}\sqrt{2 \pi}}\exp\left(\dfrac{(1 - \mu\psi_{jl})^2}{2 \psi_{jl} \mu}\right)\right)\right] - \mathbb{E}_{q(\psi_{jl}|\cdot)}\left[\log \left(\dfrac{1}{2}e^{-\frac{1}{2}\psi_{jl}}\right)\right] \\
     =& \mathbb{E}_{q(Y|\cdot)} \left[ \log Y + \log \Big(\dfrac{1}{2 \pi}\Big) + \dfrac{Y(1 - \frac{\mu}{Y})^2}{2 \mu} - \log \Big(\dfrac{1}{2}\Big) + \dfrac{1}{2Y}\right],
\end{align*}
where the third equation holds by introducing $Y = \dfrac{1}{\psi_{jl}} \sim $ InvGaussian. The above expression can be solved by using the expectations of inverse Gaussian. 

\para{KL Divergence of Double Exponential (DE) Distributions.} 
The closed form of $\text{KL}(\text{DE}(b_1) \Vert \text{DE}(b_2))$ is 
\begin{align*}
    \text{KL}(\text{DE}(b_1) \Vert \text{DE}(b_2)) = \dfrac{b_1}{b_2} + \log \dfrac{b_2}{b_1} - 1. 
\end{align*}

\para{KL Divergence of Dirichlet distributions.} 
The closed form of $\text{KL}(\text{Dir}(\bm \alpha_1) \Vert \text{Dir}(\bm \alpha_2))$ is 
\begin{align*}
    &\text{KL}(\text{Dir}(\bm \alpha_1) \Vert \text{Dir}(\bm \alpha_2)) \\
    =& \log \dfrac{\Gamma\left(\sum_{i=1}^k \alpha_{1i}\right)}{\Gamma\left(\sum_{i=1}^k \alpha_{2i}\right)} + \sum_{i=1}^k \log \dfrac{\Gamma(\alpha_{2i})}{\Gamma(\alpha_{1i})} + \sum_{i=1}^k (\alpha_{1i} - \alpha_{2i}) \left[ \psi(\alpha_{1i}) - \psi \left(\sum_{i=1}^k \alpha_{1i} \right) \right]
\end{align*}

\section{Global--Local Shrinkage Priors} \label{sec: gl-priors}

\begin{figure*}[t]
    \centering
    \includegraphics[width=0.9\textwidth]{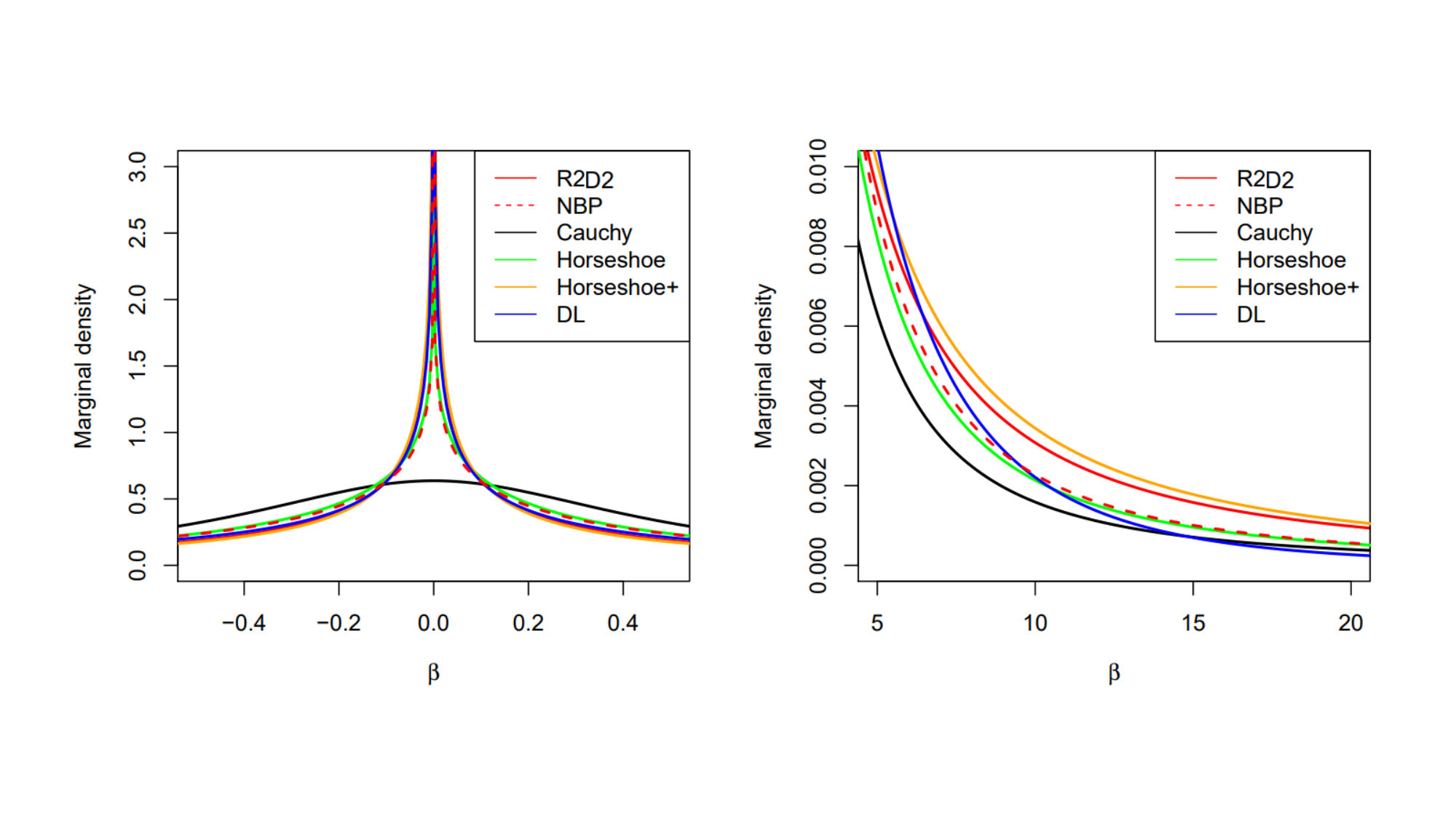}
    \caption{Marginal densities of typical global--local shrinkage priors \citep{zhang2020r2d2}. DL: Dirichlet--Laplace, NBP: normal beta prime prior.  }
    \label{fig:marginal_densities}
\end{figure*}

Figure \ref{fig:marginal_densities} presents the comparison of marginal densities of typical global--local shrinkage priors.
Table 1 in the main manuscript presents the comparisons of the concentration rate at zero and tail thickness of typical global--local shrinkage priors.
Table 1 in the main manuscript and Figure \ref{fig:marginal_densities} demonstrate that the R2D2 prior has the highest concentration rate at zero and the heaviest tail. 
The rates in Table 1 in the main text can be derived from the density functions of the global--local shrinkage priors, as given in the following. 

\para{The Horseshoe Prior.}
\begin{align*}
    \beta_j|\tau_j \sim \mathcal{N}(0, \tau_j^2) \text{ with } \tau \sim C^+(0, b_0)
\end{align*}

where $C^+$ is the Half-Cauchy distribution.

\para{The Horseshoe+ Prior.}
\begin{align*}
    \beta_j|\tau_j \sim \mathcal{N}(0, \tau_j^2) \text{ with } \tau_j | \lambda, \eta_j \sim C^+(0, \lambda\eta_j), \ \eta_j\sim C^+(0, 1).
\end{align*}


\para{The Spike-and-Slab Prior. } We adopt a spike-and-slab prior as
\begin{align*}
    \beta_j \sim \lambda \mathcal{N} (0, \sigma^2_{1, n}) + (1 - \lambda) \mathcal{N} (0, \sigma^2_{0, n}), 
\end{align*}
where $\sigma_{0, n}$ is set to be small and $\sigma_{1, n}$ is set to be relatively large. 

\para{The Dirichlet--Laplace Prior.}
The Dirichlet--Laplace prior \citep{bhattacharya2015dirichletlaplace} is given by
\begin{align*}
    \beta_j | \phi_j \sim \text{DE}(\phi_j), \ \ \phi_j \sim \text{Ga}(a^*, 1/2).
\end{align*}

\para{The Generalized Double Pareto Prior.}
The density of the generalized double Pareto prior is 
\begin{align*}
    \pi_{\text{GDP}}(\beta_j | \eta, \alpha) = (1 + |\beta_j|/\eta)^{-\alpha + 1} / (2 \eta / \alpha), \quad (\alpha, \eta > 0).
\end{align*}

\para{Alternative Form of the R2D2 Prior.}
The alternative form of the R2D2 prior allows for another formulation of the variational Gibbs inference paradigm, which is provided as follows,
\begin{align*}
    \beta_j \mid \sigma^2, \phi_j, \omega \sim \text{DE} (\sigma(\phi_j \omega/2)^{{1}/{2}}), \ \phi \sim \text{Dir}(a_\pi, \ldots, a_\pi), \ \omega \sim \text{BP}(a, b),
\end{align*}
where BP denotes the beta-prime distribution, DE denotes the double-exponential distribution, and Dir denotes the Dirichlet distribution. 

\section{Uncertainty Measures} \label{sec: uncertainty-measures}

%
The two uncertainty measures \citep{malinin2018DPN} for the OOD misclassification task are given as follows, 
\begin{itemize}
    \item Entropy:
    \begin{align*}
        \mathbb{H}[p(\bm \mu|\mathcal{D})] = - \int_{S^{K-1}} p(\bm \mu|\mathcal{D}) \log p(\bm \mu|\mathcal{D}) d \bm \mu,
    \end{align*}
    where $\mu_j$ is the normalized prediction score for class $j$.
    \item Maximum probability: we take the maximum predicted probability $\mathcal{P}$ from all classes as the confidence score,
    \begin{align*}
        \mathcal{P} = \max_c P(w_c|\mathcal{D}).
    \end{align*}
    where $P(w_c|\mathcal{D})$ is the predicted probability for class $c$.
\end{itemize}

\section{Evaluation Metrics} \label{sec: ev-metrics}

We summarize the evaluation metrics used in the experiments in the following.

\begin{itemize}
    \item Accuracy: the fraction of correct predictions to the total number of ground truth labels.
    \item F-1 score: The F-1 score for each class is defined as
    \begin{align*}
        \text{F-1 score} = 2 \cdot \dfrac{\text{precision} \cdot \text{recall}}{\text{precision} + \text{recall}}
    \end{align*}
    where `recall' is the fraction of correct predictions to the total number of ground truths in each class and precision is the fraction of correct predictions to the total number of predictions in each class.
    \item AUROC: the area under the receiver operating curve (ROC) which is the plot of the true positive rate (TPR or recall) against the false positive rate (FPR).
    \item AUPR: the area under the precision--recall curve. 
\end{itemize}

\end{document}